\documentclass[11pt]{article}

\usepackage[utf8]{inputenc}
\usepackage[T1]{fontenc}
\usepackage{lmodern}

\usepackage[letterpaper,margin=1in]{geometry}
\usepackage{microtype}

\usepackage{amsmath}
\usepackage{amssymb}

\usepackage{graphicx}
\graphicspath{{figures/}}
\usepackage{booktabs}
\usepackage{array}
\usepackage{tabularx}
\usepackage{multirow}
\usepackage{makecell}
\usepackage{adjustbox}
\usepackage{pdflscape}
\usepackage{longtable}
\usepackage{float}
\usepackage{placeins}
\usepackage{pifont}   

\usepackage[font=small,labelfont=bf]{caption}
\usepackage{xcolor}
\usepackage{enumitem}
\usepackage[normalem]{ulem}

\usepackage[round,authoryear]{natbib}
\usepackage{url}
\usepackage[colorlinks=true,linkcolor=black,citecolor=blue!50!black,urlcolor=blue!50!black]{hyperref}

\newcommand{\cmark}{\textcolor{green!55!black}{\ding{51}}}
\newcommand{\xmark}{\textcolor{red!70!black}{\ding{55}}}
\newcommand{\up}{$\uparrow$}
\newcommand{\dn}{$\downarrow$}
\renewcommand{\arraystretch}{1.15}

\numberwithin{figure}{section}

\usepackage{fancyhdr}
\newcommand{\headerlogo}{\includegraphics[height=13pt]{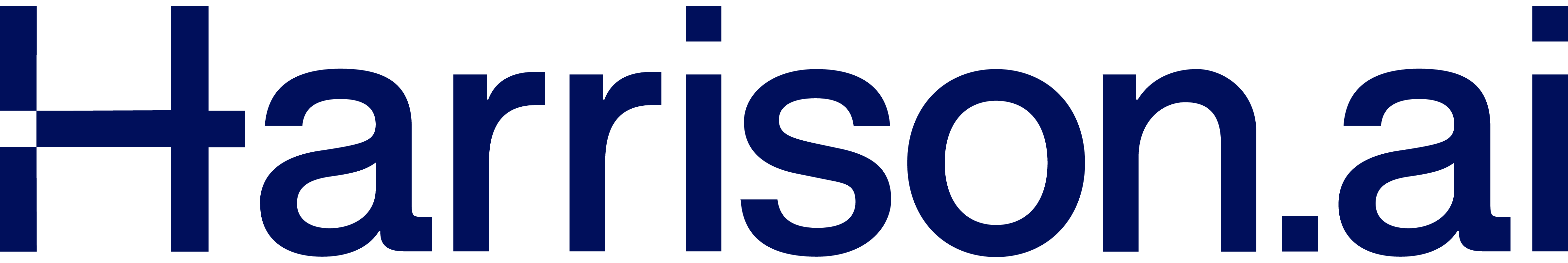}}
\fancypagestyle{hrlogo}{%
  \fancyhf{}%
  \fancyhead[L]{\headerlogo}%
  \fancyfoot[C]{\thepage}%
}
\fancypagestyle{plain}{%
  \fancyhf{}%
  \fancyhead[L]{\headerlogo}%
  \fancyfoot[C]{\thepage}%
}

\usepackage{titlesec}
\titleformat{\section}{\normalfont\Large\bfseries}{\thesection}{0.6em}{}
\titleformat{\subsection}{\normalfont\large\bfseries}{\thesubsection}{0.6em}{}
\titleformat{\subsubsection}{\normalfont\normalsize\bfseries}{\thesubsubsection}{0.6em}{}

\title{\textbf{Harrison.Rad 1.5 Technical Report}\\[4pt]
  {\large\normalfont A radiology foundation model that can draft reports from images, priors and clinical context}}

\author{
  Suneeta Mall \quad Vladimir Nekrasov \quad Ashnil Kumar \quad Sajith Karunasena \\[2pt]
  Aiden Nibali \quad Alix Bird \quad Mateo Diaz Shine \quad Jarrel Seah \\[6pt]
  \textit{Harrison.ai}
}
\date{}

\begin{document}
\maketitle

\begin{abstract}
Imaging demand is growing faster than the radiology workforce can expand, and the resulting reporting backlogs cannot be resolved through training and recruitment alone. The most direct opportunity lies in reducing the time and effort radiologists spend producing reports, a task that requires interpreting images, integrating clinical history and prior studies, and drafting structured findings. We present Harrison.Rad 1.5 (HR1.5), a radiology-specific multimodal large language model that accepts interleaved text and visual inputs and generates structured and unstructured text across the full breadth of plain-film radiology, spanning computed radiography, diagnostic x-rays for chest, musculoskeletal, abdominal, spine, and pelvic regions, and mammography. HR1.5 is trained through a three-stage pipeline: domain adaptation of a base language model on radiology reports, contrastive vision-encoder training with curriculum-based hard negatives on approximately six million image-report instances, and visual-question-answering fine-tuning on multi-turn clinical conversations. We evaluate it with a Findings-Diagnosis scoring framework that extends RadGraph-XL entity extraction with ontology-based synonym matching and polarity-contradiction detection, benchmarked on RadBench, a simulated FRCR 2B Short Case examination scored against Angoff-method thresholds, ReXGradient, and internal multi-modality datasets. HR1.5 is the only system evaluated to meet the simulated FRCR passing standard and achieves the highest accuracy on closed-format clinical questions, across anatomical regions, on internal multi-body-part and mammography report generation, and on the primary clinically-aligned score for public chest report generation. We further examine explainability and model behaviour, including question-sensitive Grad-CAM heatmaps, attention analysis, and confidence estimation, to support responsible future evaluation toward clinical use, and a proposed framework for a clinically grounded basis for assessing radiology report quality.
\end{abstract}

\footnotetext{This technical report is released for transparency and capability preview; selected sections are intended for submission to a peer-reviewed venue. This report is released for research purposes only. Harrison.rad.1.5 has not been cleared for clinical use and is not a medical device. Use must be in accordance with the Terms and Conditions.}

\setcounter{tocdepth}{3}
\clearpage
\tableofcontents
\clearpage
\section*{Introduction}
\addcontentsline{toc}{section}{Introduction}

Radiology has become the diagnostic backbone of modern medicine, and that success is now its central operational problem: imaging demand is rising faster than the workforce that reads it can grow. In the United States, imaging utilisation is projected to increase by up to 25\% over the coming decades, while radiologist supply is projected to remain constrained relative to demand \citep{Neiman2025}. In the United Kingdom, the consequences of this imbalance are already evident: the Royal College of Radiologists reports a 32\% shortfall in consultant radiologists relative to need, a deficit of nearly 2,300 radiologists \citep{RCR2025}. In 2024, imaging demand grew by 8\%, while the radiologist workforce grew by only 4.7\% \citep{RCR2025}. If these trends continue, the workforce shortfall is projected to reach 40\% by 2030 \citep{RCR2025}.

Rising imaging volumes relative to workforce growth manifests as reporting backlogs, longer turnaround times, and reader fatigue that degrades diagnostic accuracy \citep{Krupinski2010}. The radiology industry is facing a scaling problem that training and recruitment cannot solve.

Modern radiology AI solutions have already demonstrated substantial clinical value through task specific models that detect important findings, prioritise urgent studies, and assist with time-consuming measurements. These systems have no doubt improved patient care: accurate prioritisation of truly urgent cases serves a valuable function in an environment of backlogged reporting lists, however, it does not solve the underlying problem of the radiologist workforce not having capacity to read all of the scans being performed.

The greatest opportunity to address radiology's scaling challenge lies in reducing the effort and time required for radiologists to report studies, and this is a complex, multifaceted task that involves not only attention to detail in image interpretation, but also the synthesis of imaging findings with other clinical information, observation of the patient over time by comparing to prior examinations, and ultimately the integration of all of this context to consider appropriate differential diagnoses to answer a clinical question and guide the patient's treatment.

Harrison.ai's experience reflects this evolution; our current generation of AI models are trained to identify a predefined set of findings on images, and whilst they can do this with very high accuracy, they only assist the radiologist in one of the aforementioned cognitive processes required to produce a radiology report. Augmenting the radiologist on all aspects of their workflow requires an AI model that is able to integrate information the way a radiologist does by design.

Harrison.ai's next phase involves building an AI model to help solve the global radiology capacity problem from the ground up, a specialised radiology foundation model that can perform accurate detection, reason through clinical history and prior imaging, and draft a well-structured radiology report. To support the responsible development and eventual clinical translation of such systems, it is desirable that such a system also provides calibrated confidence estimates and interpretable evidence for its conclusions. We develop and evaluate each of these capabilities in the sections that follow.

These requirements have become attainable through a rapid evolution in multimodal deep learning. The Transformer \citep{Vaswani2017} and its visual counterpart, the Vision Transformer \citep{Dosovitskiy2021}, gave images and text a common modelling substrate; large-scale contrastive image-text pre-training (CLIP, \citealp{Radford2021}; SigLIP~2, \citealp{Tschannen2025}) showed that aligning the two yields transferable, open-vocabulary recognition. A subsequent line of work connected pre-trained vision encoders to large language models such as Flamingo \citep{Alayrac2022}, BLIP-2 \citep{LiBLIP2023}, and the visual instruction tuning of LLaVA \citep{Liu2023}, turning static classifiers into instruction-following systems that reason over interleaved images and text, a capability brought to general audiences by GPT-4 with vision \citep{OpenAI2023}. A vision encoder aligned to a generative language model, and instruction-tuned on multimodal dialogue is what makes a generalist, natively integrable radiology assistant feasible.

The same framework has been adapted to radiology. Contrastive alignment alone reached expert-level zero-shot detection from unannotated reports \citep[CheXzero;][]{Tiu2022}, and biomedical-scale pre-training improved transfer \citep[BiomedCLIP;][]{Zhang2023}. Instruction-tuned biomedical assistants followed \citep[LLaVA-Med;][]{LiLLaVAMed2023}, alongside increasingly generalist radiology systems: RadFM extended generative pre-training to web-scale medical data \citep{Wu2023}; ELIXR grafted an X-ray encoder onto a fixed LLM for broad chest-X-ray tasks \citep{Xu2023}; CheXagent built an instruction-tuned foundation model and benchmark for chest-X-ray interpretation \citep{Chen2024}; and MAIRA-1/MAIRA-2 advanced grounded chest-X-ray report generation \citep{Hyland2023,Bannur2024}. General-purpose medical multimodal models such as Med-Gemini \citep{Saab2024,Yang2024} and the openly released MedGemma \citep{Sellergren2025} demonstrated strong cross-domain capability, and fully open foundation models for chest radiography have since appeared \citep{Wang2025}. Figure~1.1 situates these systems by year of release, modality, and access level. Yet most of these efforts remain confined to a single modality (overwhelmingly the chest radiograph), do not span the full range of plain-film studies encountered in practice, and are rarely evaluated against the standardised examinations used to certify human radiologists.

\begin{figure}[t]
  \centering
  \includegraphics[width=0.92\linewidth]{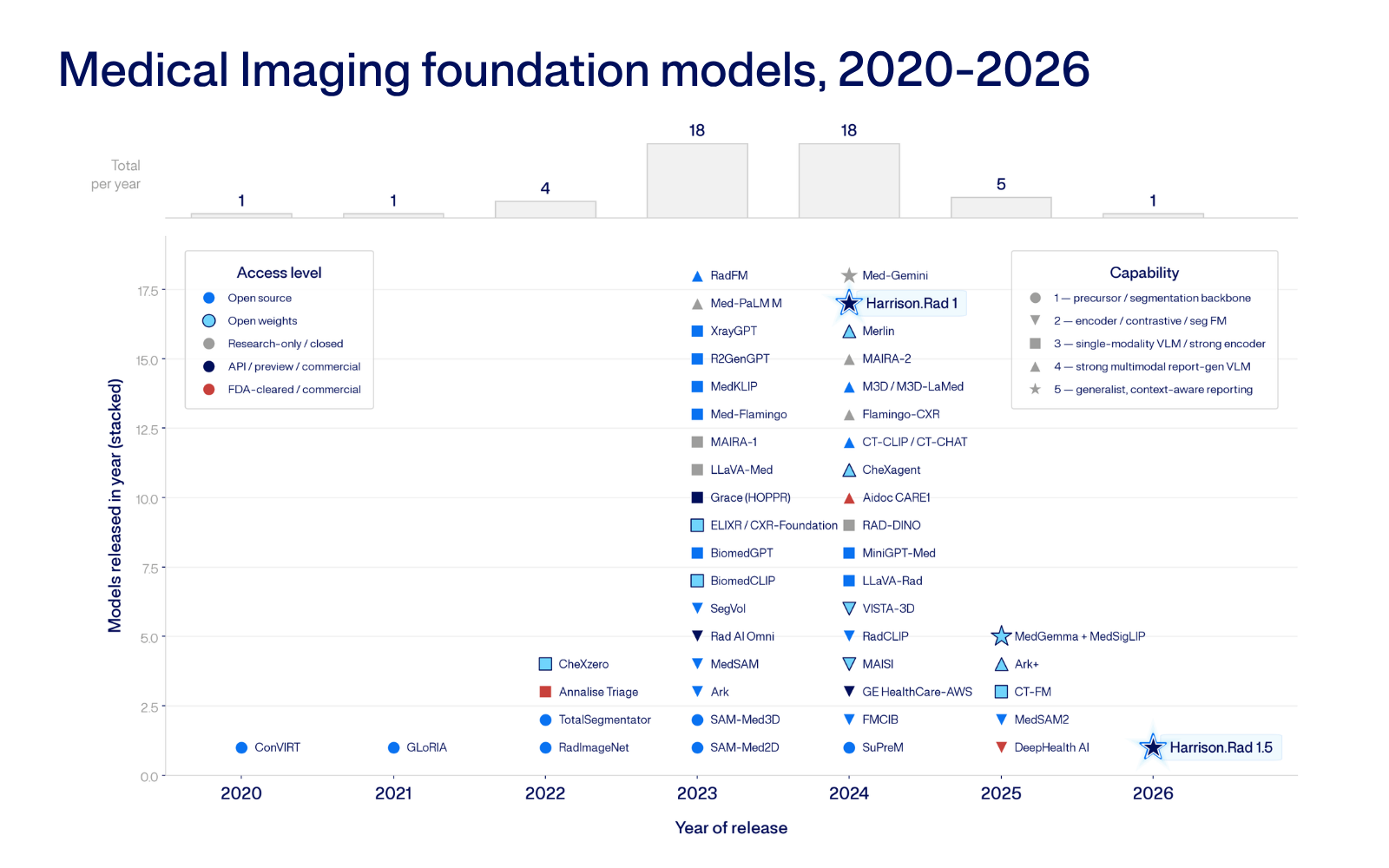}
  \caption*{\textbf{Figure 1.1.} The radiology foundation-model landscape, 2020--2026. Models are positioned by year of release, with the bars showing the total released per year; marker colour denotes access level (open source, open weights, research/closed, API/preview, FDA-cleared/commercial) and marker shape denotes capability tier, from precursor segmentation and backbone encoders (1), through contrastive or single-modality vision-language models (2--3), to strong multimodal report-generation VLMs (4) and generalist, context-aware reporting systems (5). The field is heavily concentrated on the chest radiograph and on single-modality models; the Harrison.Rad~1 and Harrison.Rad~1.5 models occupy the generalist, context-aware reporting tier.}
\end{figure}

This evaluation gap exposes a deeper limitation: open and general-purpose models tend to fail on genuinely complex cases because they lack the dense domain knowledge radiological reasoning requires. The point is concrete: on a simulation of the Fellowship of the Royal College of Radiologists (FRCR) 2B Rapids examination, which only 40--59\% of human candidates pass on first attempt, Harrison.Rad~1 scored 85.67\% (51.4/60), while every other model evaluated scored below 50\% \citep{Harrison2024}. This advantage has since been borne out independently: in a peer-reviewed multi-reader study in the \textit{American Journal of Roentgenology} involving radiologists at Stanford, Mass General Brigham, and Seoul National University Hospital, Harrison.Rad~1 was the preferred model for every reader, selected between 39.2\% and 66.5\% of the time, far ahead of the least-favoured model, Gemini~2.5 (5.7--9.4\%). It also recorded the lowest hallucination rate of any model evaluated, at 5.7\%, compared with 10.8\% for Gemini~2.5, 19.3\% for MAIRA-2, and 53.8\% for MedRAX \citep{Hong2026}. In the 2025 Healthcare AI Challenge \citep{HealthcareAIChallenge2025}, an independent evaluation conducted by the Mass General Brigham AI Arena and the American College of Radiology comprising 117 cases, 113 participants, and 2,840 evaluations, radiology reports drafted by Harrison.Rad~1 were rated clinically acceptable in 65.4\% of evaluations, compared with 79.6\% for radiologist-authored reports, indicating that the model approaches but does not yet match expert reporting quality. Capability of this kind is earned from large-scale, clinically curated imaging and reporting, not general web data. At the same time, we believe the competitive landscape should remain open enough for the radiology foundation-model ecosystem to evolve collectively, through shared benchmarks, comparable protocols, and accessible models, within the bounds of patient privacy and intellectual property. It is in this spirit that Harrison.ai makes Harrison.Rad~1 and Harrison.Rad~1.5 available for research and capability evaluation, allowing the community to probe, compare, and build on them. Currently, however, neither model is cleared or approved for clinical or diagnostic use in any jurisdiction, and must not be used to inform or drive patient care.

We present Harrison.Rad 1.5 (HR1.5), an incremental update to our radiology-specific multimodal large language model Harrison.Rad~1 (HR1) that refines its training and evaluation while retaining the same overall design. Like its predecessor, HR1.5 accepts interleaved text and visual inputs and generates structured and unstructured text outputs across the full breadth of computed radiography, diagnostic x-rays for chest, musculoskeletal, abdominal, spine, and pelvic regions, and mammography. HR1.5 follows a three-stage training pipeline: domain adaptation of a base LLM on radiology reports, contrastive vision-encoder training with curriculum-based hard negatives on approximately 6 million image-report instances, and visual-question-answering fine-tuning on crafted multi-turn clinical conversations. We evaluate it with a Findings-Diagnosis scoring framework that extends RadGraph-XL entity extraction \citep{Jain2021,Delbrouck2024} with ontology-based synonym matching and polarity-contradiction detection, benchmarked on RadBench, a simulated FRCR 2B Short Case examination scored against Angoff-method thresholds, ReXGradient \citep{Zhang2025}, and internal multi-modality datasets. We further examine model behaviour and explainability to support responsible future evaluation toward clinical use, by adapting established techniques: question-sensitive Grad-CAM heatmaps, attention analysis, and confidence estimation.
\section{Methodology}
\label{sec:method}

This section describes how HR1.5 is built and deployed: the training methodology that produces the model (\S\ref{sec:training}), the data engineering that supplies its learning signal (\S\ref{sec:data}), and the serving stack that delivers it at inference (\S\ref{sec:serving}). The emphasis throughout is on the principles behind each choice rather than on implementation specifics.

\subsection{Training Methodology}
\label{sec:training}

HR1.5 is trained through a multi-stage pipeline, an approach that has become standard for large vision--language foundation models. The now conventional strategy first establishes competence within each modality and then aligns them; a language model is adapted to the target domain, a vision encoder is aligned to text through large-scale contrastive learning in the manner of CLIP \citep{Radford2021} and SigLIP \citep{Tschannen2025}, and the two are finally connected and tuned to follow instructions over interleaved image and text inputs. This approach, pioneered by Flamingo, introduced gated cross-attention to fuse a frozen vision encoder with a language model over interleaved sequences \citep{Alayrac2022}; was made more efficient by BLIP-2's lightweight querying bridge between frozen image encoders and LLMs \citep{LiBLIP2023}, and popularised as visual instruction tuning by LLaVA \citep{Liu2023}. This method has been scaled by recent open multimodal systems such as Qwen2-VL \citep{Wang2024} and PaliGemma \citep{Beyer2024}. Staging the training in this way optimises one objective at a time before the components are trained together, e.g.\ the language model is first adapted to the target domain, the vision encoder is then aligned to text through contrastive learning, and only once these are in place are the two wired for harder joint objective and tuning to follow instructions over interleaved image and text inputs. Decoupling the stages in this way improves both training stability and sample efficiency.

This paradigm transfers naturally to radiology, and several radiology foundation models adopt a comparable multi-stage formulation; CheXagent \citep{Chen2024} and CheXOne \citep{Zhang2026}, for example, both report a multi-stage training pipeline for chest-X-ray interpretation. HR1.5 follows the same broad philosophy; several stages of pre-training with and without contrastive objectives, precede one or more fine-tuning stages that confer instruction-style, image-grounded question-answering ability. Throughout, and in keeping with the overarching goal of a clinically capable radiology assistant, the learning objectives are chosen to be predominantly radiology-relevant, so that model capacity is spent on the distinctions that matter for image interpretation and reporting rather than on general-purpose breadth.

Figure~1.2 summarises this pipeline at a conceptual level: the progression from a domain-adapted language model, through contrastive vision--language alignment, to instruction fine-tuning for visual question answering, with the dominant input modality indicated at each stage.

\begin{figure}[t]
  \centering
  \includegraphics[width=0.95\linewidth]{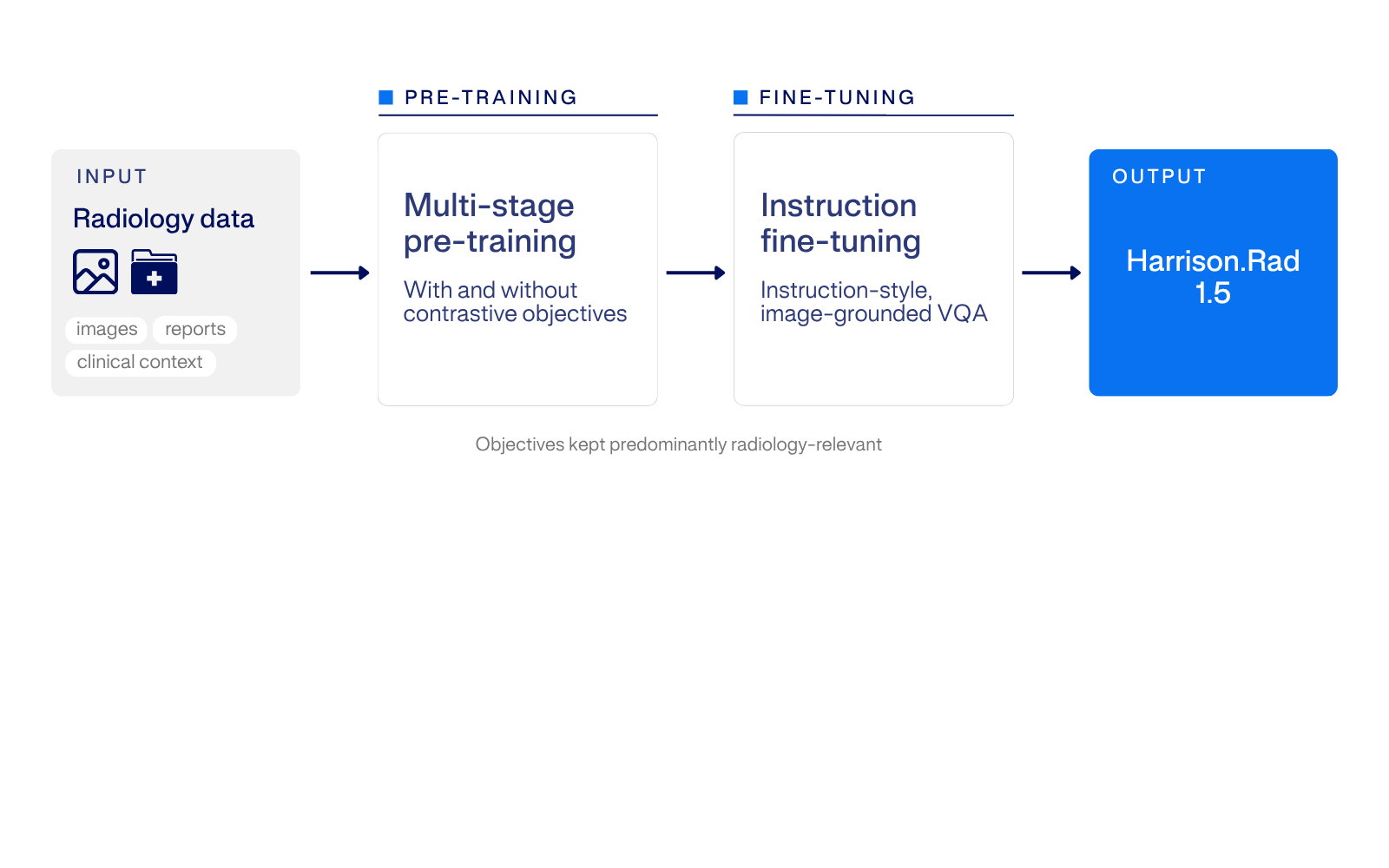}
  \caption*{\textbf{Figure 1.2.} Conceptual overview of the HR1.5 training pipeline. From radiology data (images, reports, and clinical context), training proceeds through multi-stage pre-training (with and without contrastive objectives) followed by instruction fine-tuning for image-grounded visual question answering, yielding Harrison.Rad~1.5. Objectives are kept predominantly radiology-relevant; internal architecture, stage count, and scale are intentionally omitted.}
\end{figure}

One design choice worth highlighting relates to image captioning, a well-established auxiliary objective in vision--language model training that has demonstrated clear value across general-domain systems. In our setting, however, ablations found a dedicated captioning task to be ineffective for radiology foundation-model training, neither improving nor degrading downstream performance. HR1.5 therefore omits a separate captioning objective and proceeds directly from contrastive alignment to instruction fine-tuning.

\subsubsection{Scaling Training to B200-Class Hardware}

Whereas HR1 was trained on a large fleet of NVIDIA A100 accelerators, HR1.5 was trained at scale on NVIDIA B200-class hardware. The newer accelerators offer a large step-up in compute and memory bandwidth, which moved the bottleneck off compute and onto the surrounding data and memory machinery; realising their throughput meant re-architecting that machinery rather than simply rescaling it.

Two changes mattered most. We re-architected data loading so that each training rank reads only the shard it needs, rather than every worker materialising the full dataset, removing the redundant I/O that otherwise starved the system. And we restructured on-disk storage with lazy, per-rank caching so that host memory ceased to be the binding constraint. Together, these kept the B200s consistently fed.

\subsection{Data Engineering for Training}
\label{sec:data}

The capability of HR1.5 rests as much on how its training data is engineered as on the model architecture itself. Rather than treating data as a fixed resource to be scraped and lightly augmented, we approach it as a design problem: the learning signal is constructed from the structure of the clinical task, so that each example teaches a distinction a radiologist would consider meaningful. This first-principles, mechanism-design view of crafted data (designing examples around the real-world mechanism that generates them, rather than around incidental surface statistics) guides both data workstreams that feed training.

The first workstream concerns contrastive learning, where the difficulty and informativeness of negative examples largely determine what the vision encoder learns. Building on the established observation that hard negatives sharpen fine-grained representations \citep[e.g.\ NegCLIP;][]{Yuksekgonul2023}, we extend hard-negative construction in both breadth and depth. \textit{Breadth} refers to the range of clinically meaningful axes along which a negative can differ from a true description, so that the encoder must attend to the clinically decisive feature rather than to surface wording. \textit{Depth} refers to graded difficulty: negatives follow a curriculum that begins with easily separable contrasts and progressively introduces harder, more confusable ones as the encoder improves. The aim is a representation that distinguishes descriptions that read alike but differ clinically, precisely where lexical shortcuts fail.

The second workstream concerns instruction data, which we craft anchored in real clinical context. Genuine case material serves as the source from which diagnostic questions and multi-turn exchanges are derived, spanning the range of tasks a radiologist performs: overall assessment, description of specific findings, verification of a suspected finding, differential reasoning, anatomical localisation, and comparison against prior studies. Coverage is shaped deliberately rather than left to the empirical frequency of conditions (under-represented pathologies and challenging presentations are oversampled, and example difficulty is controlled) so that the model is not biased toward common findings at the expense of the rare-but-important ones. The guiding principle throughout is fidelity to the mechanism that produces real radiological reasoning, so that scale never comes at the cost of clinical realism.

Figure~1.3 illustrates this philosophy schematically: a single de-identified clinical source is transformed, along one path, into breadth-and-depth hard negatives for contrastive alignment and, along the other, into coverage-controlled crafted dialogues for instruction tuning.

\begin{figure}[t]
  \centering
  \includegraphics[width=0.82\linewidth]{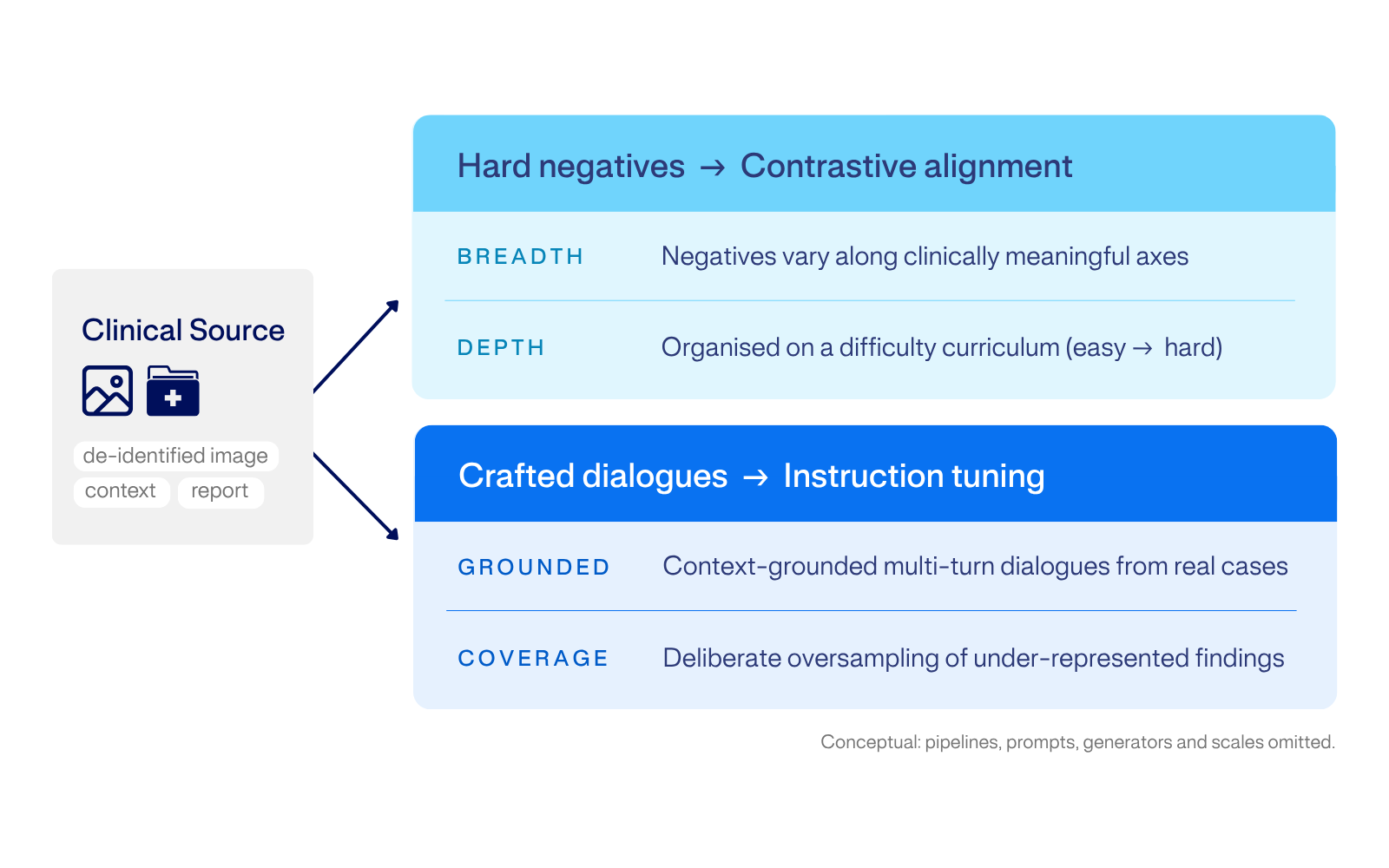}
  \caption*{\textbf{Figure 1.3.} Data engineering for HR1.5 training, shown conceptually. From a de-identified clinical source, two engineered learning signals are produced: (top) hard negatives that vary along clinically meaningful axes (breadth) under a difficulty curriculum (depth), for contrastive alignment; (bottom) context-grounded crafted dialogues derived from real cases, with deliberate oversampling of under-represented findings, for instruction tuning. Specific pipelines, prompts, generator models, and dataset scales are intentionally not shown.}
\end{figure}

All data used for training and evaluation are de-identified following strict institutional and regulatory guidelines, so that no personal health information is present at any stage of the pipeline.

In aggregate, HR1.5 was trained on approximately 6.5 million studies, yielding in the order of 18 million conversations spanning a broad range of clinical tasks.

\subsection{Serving}
\label{sec:serving}

Because HR1.5 and its predecessor HR1 use a custom architecture, we built a bespoke serving integration on top of vLLM \citep{Kwon2023}, the high-throughput, low-latency inference engine for large language models.

At inference, we apply methods that curtail hallucination and keep answers grounded in the clinically meaningful content of the input. Our approach is inspired by contrastive decoding \citep{LiContrastive2023,Leng2024}, which reduces bias and hallucination by contrasting output distributions, but is developed internally and adapted to our objective of clinically faithful generation. Together with the served model, this constitutes the \textbf{HR1.5 Core API}.

We extend the Core API through an agentic framework for higher-level tasks such as report generation and chat-completion/question-answer style interaction, using an architecture analogous to a system of experts. Within this workflow the constituent models are consulted according to the clinical task at hand, other harrison.ai models may contribute where appropriate, however HR1.5 is always consulted. This agentic configuration represents the \textbf{HR1.5+} system reported in the evaluations that follow.
\section{Quantitative Results}
\label{sec:quant}

We evaluate HR1.5 across a comprehensive suite of benchmarks spanning open-ended and closed-format visual question answering, structured report generation, and clinical examination performance. Evaluations are reported across all models compared (HR1.5, HR1, CheXagent \citep{Chen2024}, CheXOne \citep{Zhang2026}, MedGemma \citep{Sellergren2025}, GPT-5.4 \citep{OpenAI2026}, Gemini-3-Flash-Preview \citep{Google2025}, Claude Opus 4.7 \citep{Anthropic2026}) wherever applicable, with results broken down by body part and question type. We additionally evaluated several other open-source models, but on manual review deemed them unsuitable for the clinical tasks assessed here; these models were therefore removed from the comprehensive review.

The open-weight baselines were served behind OpenAI-compatible endpoints (vLLM, except CheXagent which runs locally via HuggingFace Transformers) and queried greedily (temperature 0): HR1 and HR1.5 used a 300-token generation cap with \texttt{top\_k=1}, a frequency penalty of 1.1, and JSON-schema-constrained (guided) decoding for closed questions; MedGemma (4B/27B) used a 500-token cap; CheXOne used a 300-token cap in instruct mode and 1024 tokens in reasoning mode, both with a repetition penalty of 1.1; and CheXagent generated up to 300 new tokens in bfloat16 using its shipped \texttt{generation\_config}. For the proprietary API models, decoding temperature was set to 0 wherever the model accepts it (GPT-5.4 and Gemini-3-Flash-Preview) and left at the provider default otherwise (Claude Opus 4.7, which does not respect a temperature parameter). Generation was capped at 512 output tokens for GPT-5.4 and Gemini-3-Flash-Preview and at 100 tokens for Claude Opus 4.7; GPT-5.4 was queried through the OpenAI Responses API with reasoning effort ``none'', and Gemini-3-Flash-Preview through its OpenAI-compatible endpoint with reasoning effort ``medium''. All remaining sampling parameters were left at their provider or server defaults.\nocite{Meta2025}

\subsection{Evaluation}
\label{sec:evaluation}

Our evaluation is designed to mirror both how radiologists are certified and how they work day-to-day. At its core, is a mock simulation of the Royal College of Radiologists (RCR) FRCR examination; the assessment used to certify practising radiologists. This simulation reconstructs the exam in both its legacy Rapid Reporting and current Short Case forms (Section~\ref{sec:frcr}). The legacy FRCR Rapids component (retired in 2025) presented 30 plain radiographs per exam sheet, roughly half of which were normal, to be reported within 35 minutes. For each film the candidate recorded only ``normal'' or the key diagnosis, rewarding fast, accurate triage and penalising over-calling. With a pass mark of 54 of 60, a score of 2 per question was rewarded if diagnosis was identified correctly, otherwise 0. The current FRCR 2B Short Cases format (the new standard since 2025) replaced Rapid Reporting, here we present 25 abnormal cases drawn mainly from chest (50--60\%), musculoskeletal (40--50\%) and abdominal (up to 4\%) radiographs, each case scored out of up to five marks (total score 125). Around this we measure open- and closed-format question answering on plain-film X-ray interpretation tasks, so that both free-text reporting and discrete clinical decisions are captured.

The benchmarks span public and internal (held out set) sources across a range of body parts and question types: clinician-authored cases (\href{https://github.com/harrison-ai/radbench}{RadBench}), the FRCR examination simulations, the large-scale public chest-X-ray dataset RexGradient \citep{Zhang2025}, an internal held-out clinical question set built for broad body-part coverage over a mixed set of medium-to-hard clinical tasks (RadCoverage-VQA), and mammography (\href{https://www.cancerimagingarchive.net/collection/cbis-ddsm/}{CBIS-DDSM}). Together these probe HR1.5 from standardised examination performance through to everyday reporting across the breadth of plain-film radiology. The scoring methodology common to these evaluations is described next.

\subsubsection*{Evaluation Scoring: Findings-Diagnosis Methodology}

Our primary evaluation scoring extends RadGraph-XL entity extraction \citep{Jain2021,Delbrouck2024} into a Findings-Diagnosis framework used across RadCoverage-VQA, RexGradient, and mammography benchmarks. RadGraph-XL extracts clinical entities (anatomy, observations with polarity labels) and their relations from both predicted and ground-truth text. The Findings-Diagnosis score then applies:

\begin{itemize}[leftmargin=1.4em]
  \item \textbf{Normal/abnormal classification}: Pattern-based text analysis with negation-aware phrase matching determines whether the prediction indicates a normal or abnormal study; when text matching is ambiguous, RadGraph entity analysis checks for the presence of pathological observations as a fallback.
  \item \textbf{Diagnosis overlap scoring}: RadGraph-XL extracts clinical entities from both texts; observation tokens are filtered to remove generic descriptors (severity, laterality modifiers) and meta-terms (e.g., ``finding'', ``lesion''), then lemmatised and matched using fuzzy token matching with ontology-based synonym expansion from a radiology knowledge graph.
  \item \textbf{Polarity contradiction detection}: Cases where the prediction asserts the opposite polarity to the ground truth (e.g., definitely denying a finding that is definitely present, or vice versa) receive zero credit; this is checked bidirectionally across both prediction and reference entity sets.
\end{itemize}

This scoring is applied consistently across open-ended and closed-format evaluations. For closed-format (multiple-choice) questions, an LLM-as-a-judge approach \citep{Gu2024} validates answer correctness when exact matching is insufficient.

Comparing every metric on the same studies, scored across models, makes the deficiencies of conventional metrics concrete. These metrics track how a report is \textit{worded} far more than whether it is \textit{clinically correct}. Lexical-overlap scores (BLEU, ROUGE-L) reward n-gram agreement with the reference phrasing, so they collapse toward zero for a factually excellent report worded differently and can rank a fluent, correct paraphrase below a clumsy near-copy of the reference. Embedding- and entity-based scores (BERTScore, CheXbert-cos, RadGraph) reward semantic or entity similarity but are largely insensitive to clinical polarity and salience, so they remain high even when a report hallucinates a clinically significant finding or misses one. Composite error metrics such as RadCliQ inherit the same blind spots from their components. In short, none of these metrics reliably separates a clinically correct report from a clinically wrong one, which is why we do not treat them as indicative of clinical quality.

The Findings-Diagnosis score is introduced to close this gap: by comparing extracted findings and penalising polarity contradictions, it separates clinically correct from clinically incorrect reports far more faithfully than any surface metric. It is not, however, infallible; its normal/abnormal decision and finding overlap are computed automatically, it can still credit a report that misses a subtle finding or accept a vague but incorrect descriptor as a match. We therefore adopt the following reporting policy for the open-ended evaluations that follow: the Findings-Diagnosis score is the primary measure, and where its automated finding-matching is insufficient to adjudicate clinical correctness, most notably on the free-text report generation of RexGradient, we additionally report a Judge Findings-Diagnosis score, an LLM-as-a-judge \citep{Gu2024} adjudication of the same findings-versus-diagnosis comparison. Conventional report-generation metrics are still tabulated for consistency with the prior literature, but should be read only as context alongside these clinically aligned scores.

To make this concrete, the cases in Table~\ref{tab:illustrative} place the conventional metrics beside a clinician reference verdict (\textbf{GT acceptable} --- whether a radiologist would accept the report) and the Findings-Diagnosis (F-D) verdict with its judge-based adjudication (Judge F-D), so the automated verdicts can be read directly against the clinical reference. The true (ground-truth) answer for each is given in the first column and the predicted answer verbatim in the second; \cmark/\xmark{} indicate whether the prediction is clinically acceptable.

\begin{landscape}
\begin{table}
\centering
\caption{Illustrative per-study comparison of conventional metrics against the Findings-Diagnosis (F-D) verdict, the judge-based Judge F-D, and a clinician acceptability reference (GT acceptable).}
\label{tab:illustrative}
\scriptsize
\renewcommand{\arraystretch}{1.1}
\begin{adjustbox}{max width=\textheight}
\begin{tabular}{@{}p{4.0cm}p{6.2cm}cccccccc@{}}
\toprule
\textbf{True answer} & \textbf{Predicted answer} & \textbf{RadGraph-F1} & \textbf{CheXbert-cos} & \textbf{BERTScore} & \textbf{ROUGE-L} & \textbf{BLEU} & \textbf{GT acc.} & \textbf{F-D} & \textbf{Judge F-D} \\
\midrule
These findings suggest degenerative changes. Anterolisthesis indicates C4 is slightly displaced forward over C5. Narrowed discs and facet joint arthropathy suggest wear. Foraminal stenosis could lead to nerve compression. The left convexity is a mild postural or structural curve. These changes are common in elderly patients.
& MedGemma-27B: ``Cervical spondylosis.''
& 0.000 & 0.609 & 0.819 & 0.000 & 0.000 & \cmark & \cmark & \cmark \\
\addlinespace
The Hill-Sachs deformity is characterized by a compression fracture in the humeral head, which is consistent with recurrent anteroinferior dislocation.
& MedGemma-27B: ``Cannot be determined from this image.''
& 0.000 & 0.307 & 0.819 & 0.000 & 0.000 & \xmark & \xmark & \xmark \\
\addlinespace
Increasing ill-defined airspace consolidation in the right lung base partially obscuring the right hemidiaphragm. Diffuse peribronchial cuffing. Small right pleural effusion. Ill-defined opacity in the medial aspect of the left lung base as well. No evidence of pulmonary edema. Heart size is normal. Calcifications of the mitral annulus. Atherosclerotic calcifications in the thoracic aorta. Upper mediastinal contours are within normal limits. Old healed fractures of the posterolateral aspects of the right eighth and ninth ribs.
& HR1.5+: ``The chest X-ray demonstrates patchy airspace consolidation in the right lower lung zone, consistent with pneumonia in the clinical context of shortness of breath. Small bilateral pleural effusions are present with blunting of the costophrenic angles bilaterally. Bilateral lower zone interstitial thickening is noted. Aortic arch calcification is evident, representing age-related vascular changes. Chronic healed or healing rib fractures are present on the right side, involving multiple ribs in the mid to lower thoracic region, appearing stable without acute displacement. The cardiac silhouette is normal in size without cardiomegaly. No pulmonary edema or pneumothorax is identified. The mediastinum appears unremarkable. The lung bases are incompletely imaged, which may limit full assessment of pathology in those regions. No calcified pleural plaques are identified.''
& 0.184 & 0.624 & 0.855 & 0.199 & 0.054 & \cmark & \cmark & \cmark \\
\addlinespace
\textit{(same ground truth as above)}
& MedGemma-27B: ``The cardiomediastinal silhouette is within normal limits. The hila are unremarkable. There are diffuse bilateral interstitial opacities, predominantly in the lower lung zones, with a reticular pattern and some areas of ground-glass opacity. There is also evidence of bronchial wall thickening, particularly in the lower lobes. There are no pleural effusions or pneumothorax identified. The visualized bony structures appear unremarkable.''
& 0.017 & 0.461 & 0.843 & 0.187 & 0.048 & \xmark & \xmark & \xmark \\
\addlinespace
The bony pelvic ring is intact, but there is severe narrowing of the left hip joint space with subarticular sclerosis, subchondral cyst formation, and marginal osteophytes. Additionally, a mixed lytic and sclerotic lesion is present in the acetabulum and ilium, with expansile features, cortical breach, and a soft tissue component.
& HR1.5+: ``Mixed lytic and sclerotic lesion in the left acetabulum with cortical irregularity and protrusio, moderate bilateral hip joint space narrowing with subarticular sclerosis and marginal osteophytes, and degenerative changes in the sacroiliac joints and lower lumbar spine.''
& 0.275 & 0.843 & 0.899 & 0.345 & 0.102 & \cmark & \cmark & \cmark \\
\addlinespace
\textit{(same ground truth as above)}
& MedGemma-4B: ``Pelvic asymmetry''
& 0.000 & 0.653 & 0.812 & 0.038 & 0.000 & \xmark & \xmark & \xmark \\
\addlinespace
Mild retrolisthesis at C4/5 and C5/6. Moderate to severe disc degenerative changes at C4/5 and C5/6 with moderate disc degenerative changes at C6/7. Right C3/4 facet joint arthropathy. Left neural foramina are patent. Mild C5/6 foraminal narrowing due to uncovertebral osteophytes.
& HR1.5+: ``Mild retrolisthesis at C4/5 and anterolisthesis at C7/T1. Severe degenerative disc disease from C4/5 to C6/7, with moderate degeneration at C3/4. Bilateral facet joint arthropathy, present on the left from C3/4 to C6/7 and on the right from C2/3 to C5/6. Severe bony foraminal stenosis bilaterally at C2/3. Moderate left foraminal stenosis at C3/4 and moderate right foraminal stenosis at C3/4. Mild left foraminal stenosis at C4/5. Large osteophytes present at multiple levels. No bone lesions or fractures identified.''
& 0.259 & 0.661 & 0.871 & 0.284 & 0.117 & \cmark & \xmark & \cmark \\
\bottomrule
\end{tabular}
\end{adjustbox}
\end{table}
\end{landscape}

Several patterns illustrate why the surface metrics are not indicative of clinical quality, and why we rely on the Findings-Diagnosis family instead; throughout, the GT-acceptable column is the clinician reference against which the automated verdicts are read. First, lexical-overlap metrics (BLEU, ROUGE-L) and even RadGraph-F1 collapse to zero for a terse but clinically correct answer (\textit{``Cervical spondylosis''}), whereas F-D and Judge F-D both credit the correct diagnosis. Second, those same metrics give near-identical scores to genuinely uninformative outputs: a non-answer (\textit{``Cannot be determined from this image.''}) and a vague non-specific descriptor (\textit{``Pelvic asymmetry''}) hold BERTScore around 0.81--0.82 (comparable to fully correct reports) yet both F-D and Judge F-D correctly mark them incorrect. Third, a clinically correct report and a clinically wrong one on the \textit{same} study can sit within 0.01--0.09 BERTScore of each other (the chest and pelvis pairs above); the embedding- and entity-based scores cannot separate them, but F-D marks the report that mischaracterises the consolidation and denies the effusion, or offers only ``pelvic asymmetry'', as incorrect. Finally, where automated finding-matching is confounded by differing level labels and marks a thorough, clinically sound report incorrect (automated F-D \xmark), the judge adjudication recovers the clinical agreement (Judge F-D \cmark), the regime that motivates reporting Judge F-D on free-text report generation such as RexGradient.

\subsection{Results}
\label{sec:results}

\subsubsection{FRCR Evaluations --- Mock examination}
\label{sec:frcr}

The FRCR examinations are not publicly available. We created a simulation of the FRCR examination, using external data that was not used for model training. In building the examination, we made our best effort to follow the published examination guidelines.

All results in this section are reported as \textbf{sheets passed} (for FRCR-style exams) or per-benchmark aggregate metrics. We report sheets passed to provide a clinically meaningful measure: a model either meets the standard for a given exam sheet or it does not. However, we also report median and interquartile range (IQR; 25th--75th percentiles) of the scores to provide a balance review across series of mock examination sheets.

\paragraph{Legacy FRCR Rapids (Retired since 2025).}
The legacy FRCR 2B Rapid Reporting format consists of a mix of normal and abnormal plain-film X-ray cases. The RCR retired this format in 2025, having replaced it with the Short Case examination (Section~\ref{sec:frcr}); we retain it here as an additional point of comparison. We evaluate on 70 held-out exam sheets. Each case is assessed in two sequential parts: first, whether the study is normal or abnormal; and second, where the study is abnormal, the findings that are present. A case is scored correct only when both parts are answered correctly: the normal/abnormal call must be right and, if the case is abnormal, the findings must also be correctly identified. A fully correct case scores 2 and any other response scores 0, giving a maximum of 60 per sheet. Following the FRCR Rapids guideline, a sheet is counted as passed at a score of 54 or above. Results are reported as the number of exam sheets passed.

\begin{table}[H]
\centering
\caption{Legacy FRCR 2B Rapid Reporting (retired): per-sheet pass rate.}
\label{tab:frcr-rapids}
\begin{tabular}{@{}lc@{}}
\toprule
\textbf{Model} & \textbf{Pass Rate} \\
\midrule
HR1.5+ & 15.7\% \\
\textbf{HR1.5} & \textbf{24.3\%} \\
HR1 & 10.0\% \\
\midrule
GPT-5.4 & 0.0\% \\
Gemini-3-Flash-Preview & 0.0\% \\
Claude Opus 4.7 & 0.0\% \\
MedGemma-27B & 0.0\% \\
MedGemma-4B & 0.0\% \\
\bottomrule
\end{tabular}
\end{table}

As shown in Table~\ref{tab:frcr-rapids}, HR1.5 leads with 24.3\%, followed by HR1.5+ at 15.7\%, and HR1 at 10.0\%. None of the general-purpose models nor the medical-domain MedGemma \citep{Sellergren2025} variants pass a single sheet, underscoring the difficulty of the legacy FRCR Rapids format and the advantage of radiology-specific training. We attribute the slight regression of HR1.5+ on FRCR 2B Rapids to its greater tendency to produce detailed, characterising responses, which are penalised on normal cases. The same effect is visible in Table~\ref{tab:closed-qtype}, where HR1.5+ also regresses slightly on the closed normal/abnormal question type. This is particularly consequential for FRCR 2B Rapids, where roughly half of the cases in each sheet are normal.

\paragraph{FRCR 2B Short Cases (New Standard, 2025).}
The FRCR 2B Short Case examination represents the new RCR standard since 2025, consisting exclusively of abnormal cases requiring structured report writing. We evaluate on 16 held-out exam sheets, each containing 25 abnormal X-ray cases with a body-part distribution mirroring the official RCR specification (approximately 56\% chest, with the remainder across MSK upper/lower limb, spine, pelvis, and abdomen).

The model is asked to describe the findings for each case. Responses are scored on a 0--5 rubric by an LLM-as-a-judge \citep{Gu2024}, with diagnosis as the primary scoring axis aligned to official FRCR 2B marking criteria. Each sheet yields a cumulative score out of 125.

Following RCR guidance, we estimated the Angoff cut-off \citep{Angoff1971} for each exam sheet. These cut-offs were spot-checked and verified by internal expert radiologists. The human expert cut-off came in significantly lower ($\sim$50\%) than the estimated Angoff cut-off, indicating that we are holding our models to a deliberately high standard. Results are reported as sheets passed against the Angoff cut-off.

\begin{table}[H]
\centering
\caption{FRCR 2B Short Case examination: mean and median sheet scores (out of 125) along with its 95\% bootstrap confidence interval.}
\label{tab:frcr-short-scores}
\begin{tabular}{@{}lcc@{}}
\toprule
\textbf{Model} & \textbf{Median (/125)} & \textbf{Mean (95\% CI)} \\
\midrule
\textbf{HR1.5+} & \textbf{86.5} & \textbf{74.8} (61.4--87.2) \\
HR1.5 & 73.5 & 63.5 (52.6--73.5) \\
HR1 & 71.0 & 60.7 (48.4--72.4) \\
\midrule
GPT-5.4 & 44.0 & 44.6 (39.2--50.0) \\
Gemini-3-Flash-Preview & 38.0 & 36.9 (32.7--40.8) \\
Claude Opus 4.7 & 23.0 & 23.6 (17.8--29.8) \\
MedGemma-27B & 18.0 & 21.7 (15.6--28.4) \\
MedGemma-4B & 11.0 & 17.7 (11.4--24.6) \\
LLaMa-4-Scout & 12.0 & 14.5 (9.7--19.6) \\
\bottomrule
\end{tabular}
\end{table}

Table~\ref{tab:frcr-short-scores} presents the FRCR 2B Short Case results. The Angoff cut-off ranges from 69.3 to 77.3 across sheets (mean 73.2). HR1.5+ and HR1.5 exceed the mean Angoff cut-off (median of 86.5 and 73.5, respectively), with HR1 (median 71.0) falling near the cut-off boundary. All external models scored well below the passing threshold.

Scoring each sheet against its own set cut-off mark and counting a sheet as passed only when its cumulative score meets that cut-off, the per-sheet pass rate is as follows:

\begin{table}[H]
\centering
\caption{FRCR 2B Short Case examination: per-sheet pass rate against the Angoff cut-off.}
\label{tab:frcr-short-pass}
\begin{tabular}{@{}lc@{}}
\toprule
\textbf{Model} & \textbf{Pass Rate} \\
\midrule
\textbf{HR1.5+} & \textbf{62.5\%} \\
HR1.5 & 50\% \\
HR1 & 50\% \\
\midrule
GPT-5.4 & 0.0\% \\
Gemini-3-Flash-Preview & 0.0\% \\
Claude Opus 4.7 & 0.0\% \\
MedGemma-27B & 0.0\% \\
MedGemma-4B & 0.0\% \\
\bottomrule
\end{tabular}
\end{table}

As shown in Table~\ref{tab:frcr-short-pass}, HR1.5+ leads with 62\% rate, and HR1 at 50.0\%. None of the general-purpose models nor the medical-domain MedGemma \citep{Sellergren2025} variants pass a single sheet, underscoring the difficulty of the FRCR 2B Short Cases format and the advantage of radiology-specific training.

\subsubsection{Closed Question Datasets}
\label{sec:closed}

To complement publicly available benchmarks, we curated closed evaluation datasets targeting specific pathology categories and clinical scenarios. These datasets were generated using a structured pipeline: an LLM produces clinically plausible multiple-choice questions and distractors from ground-truth case data, and generated questions are subsequently reviewed and validated by expert radiologists.

Categories covered include abdominal pathologies, airspace opacification, chest lines and tubes, masses and bone lesions, C-spine levels, and body-part positioning, among others.

We report accuracy broken down by question type across all closed datasets (RadBench-Closed, ReXGradient-Closed, FRCR-Closed, Pathology-Closed), pooling items of the same type regardless of source dataset. This avoids aggregation artefacts from dataset-size imbalance. We restrict to question types with sufficient sample size ($n>500$) and note the random baseline for each, which varies with the number of answer options.

\begin{table}[H]
\centering
\caption{Closed-format accuracy by question type, along with the 95\% Wilson score confidence intervals.}
\label{tab:closed-qtype}
\small
\begin{adjustbox}{max width=\textwidth}
\begin{tabular}{@{}lcccc@{}}
\toprule
\textbf{Model} & \makecell{\textbf{Diagnosis}\\\textbf{(n=13390)}} & \makecell{\textbf{Finding}\\\textbf{(n=1443)}} & \makecell{\textbf{Normal/Abn}\\\textbf{(n=1053)}} & \makecell{\textbf{Laterality}\\\textbf{(n=534)}} \\
\midrule
Random baseline & 25.0\% & 50.0\% & 50.0\% & 33.3\% \\
HR1.5 & \makecell{\textbf{82.0\%}\\\textbf{(81.3--82.6)}} & \makecell{\textbf{81.2\%}\\\textbf{(79.1--83.1)}} & \makecell{\textbf{82.8\%}\\\textbf{(80.4--85.0)}} & \makecell{85.4\%\\(82.2--88.1)} \\
HR1.5+ & \makecell{78.0\%\\(77.3--78.7)} & \makecell{77.4\%\\(75.2--79.5)} & \makecell{78.3\%\\(75.7--80.7)} & \makecell{85.5\%\\(82.3--88.2)} \\
HR1 & \makecell{75.3\%\\(74.6--76.0)} & \makecell{78.0\%\\(75.8--80.1)} & \makecell{81.9\%\\(79.5--84.1)} & \makecell{68.7\%\\(64.6--72.5)} \\
GPT-5.4 & \makecell{77.6\%\\(76.9--78.3)} & \makecell{73.0\%\\(70.7--75.2)} & \makecell{70.6\%\\(67.8--73.3)} & \makecell{\textbf{88.2\%}\\\textbf{(85.2--90.7)}} \\
Gemini-3-Flash-Preview & \makecell{67.9\%\\(67.1--68.7)} & \makecell{77.1\%\\(74.9--79.2)} & \makecell{66.2\%\\(63.3--69.0)} & \makecell{80.0\%\\(76.4--83.2)} \\
Claude Opus 4.7 & \makecell{59.9\%\\(59.1--60.7)} & \makecell{58.1\%\\(55.5--60.6)} & \makecell{56.6\%\\(53.6--59.6)} & \makecell{77.3\%\\(73.6--80.7)} \\
MedGemma-27B & \makecell{79.3\%\\(78.6--80.0)} & \makecell{51.3\%\\(48.7--53.9)} & \makecell{63.8\%\\(60.9--66.6)} & \makecell{50.2\%\\(46.0--54.4)} \\
MedGemma-4B & \makecell{75.3\%\\(74.6--76.0)} & \makecell{76.5\%\\(74.2--78.6)} & \makecell{53.7\%\\(50.7--56.7)} & \makecell{58.2\%\\(54.0--62.3)} \\
LLaMa-4-Scout & \makecell{46.0\%\\(45.2--46.8)} & \makecell{55.3\%\\(52.7--57.8)} & \makecell{50.8\%\\(47.8--53.8)} & \makecell{66.3\%\\(62.2--70.2)} \\
\bottomrule
\end{tabular}
\end{adjustbox}
\end{table}

Table~\ref{tab:closed-qtype} shows per-question-type accuracy. HR1.5 leads in Diagnosis (82.0\%), Finding (81.2\%), and Normal/Abnormal (82.8\%). GPT-5.4 achieves the highest Laterality accuracy (88.2\%). All models substantially exceed random baselines, with the largest margins on Diagnosis where chance is 25\%.

Table~\ref{tab:closed-bodypart} presents overall closed-question accuracy grouped by body-part category. Body parts are mapped to four groups: CXR (chest/thorax), MSK (musculoskeletal, including spine, extremities, pelvis, head, and dental), Abdomen, and Other. ReXGradient items are assigned to CXR as the dataset contains only chest X-rays.

\begin{table}[H]
\centering
\caption{Closed-format accuracy by body-part group.}
\label{tab:closed-bodypart}
\small
\begin{adjustbox}{max width=\textwidth}
\begin{tabular}{@{}lcccc@{}}
\toprule
\textbf{Model} & \makecell{\textbf{CXR}\\\textbf{(n=12103)}} & \makecell{\textbf{MSK}\\\textbf{(n=3768)}} & \makecell{\textbf{Abdomen}\\\textbf{(n=143)}} & \makecell{\textbf{Other}\\\textbf{(n=826)}} \\
\midrule
HR1.5 & \makecell{\textbf{79.0\%}\\\textbf{(78.3--79.7)}} & \makecell{\textbf{88.8\%}\\\textbf{(87.8--89.8)}} & \makecell{80.4\%\\(73.1--86.1)} & \makecell{\textbf{80.3\%}\\\textbf{(77.4--82.9)}} \\
HR1.5+ & \makecell{76.2\%\\(75.4--77.0)} & \makecell{82.2\%\\(80.9--83.4)} & \makecell{74.8\%\\(67.1--81.2)} & \makecell{69.6\%\\(66.4--72.6)} \\
HR1 & \makecell{73.1\%\\(72.3--73.9)} & \makecell{80.0\%\\(78.7--81.2)} & \makecell{74.1\%\\(66.4--80.6)} & \makecell{73.5\%\\(70.4--76.4)} \\
GPT-5.4 & \makecell{77.5\%\\(76.7--78.2)} & \makecell{76.6\%\\(75.2--77.9)} & \makecell{\textbf{84.6\%}\\\textbf{(77.8--89.6)}} & \makecell{57.3\%\\(53.9--60.6)} \\
Gemini-3-Flash-Preview & \makecell{67.9\%\\(67.1--68.7)} & \makecell{68.3\%\\(66.8--69.8)} & \makecell{79.0\%\\(71.6--84.9)} & \makecell{71.2\%\\(68.0--74.2)} \\
Claude Opus 4.7 & \makecell{59.5\%\\(58.6--60.4)} & \makecell{58.7\%\\(57.1--60.3)} & \makecell{60.8\%\\(52.6--68.4)} & \makecell{65.0\%\\(61.7--68.2)} \\
MedGemma-27B & \makecell{65.6\%\\(64.7--66.4)} & \makecell{52.8\%\\(51.2--54.4)} & \makecell{44.4\%\\(36.5--52.6)} & \makecell{61.9\%\\(58.5--65.1)} \\
MedGemma-4B & \makecell{64.6\%\\(63.7--65.4)} & \makecell{68.1\%\\(66.6--69.6)} & \makecell{60.5\%\\(52.3--68.1)} & \makecell{50.8\%\\(47.4--54.2)} \\
CheXOne-Reasoning & \makecell{77.6\%\\(76.8--78.3)} & -- & -- & -- \\
CheXOne-Instruct & \makecell{72.7\%\\(71.9--73.5)} & -- & -- & -- \\
CheXagent & \makecell{34.7\%\\(33.9--35.6)} & -- & -- & -- \\
LLaMa-4-Scout & \makecell{44.5\%\\(43.6--45.4)} & \makecell{55.0\%\\(53.4--56.6)} & \makecell{58.7\%\\(50.5--66.4)} & \makecell{56.3\%\\(52.9--59.6)} \\
\bottomrule
\end{tabular}
\end{adjustbox}
\end{table}

HR1.5 leads in CXR (79.0\%), MSK (88.8\%), and Other (80.3\%). GPT-5.4 achieves the highest Abdomen accuracy (84.6\%), though the Abdomen subset is small (n=143). CheXOne-Reasoning reaches 77.6\% on CXR, competitive with GPT-5.4 but below HR1.5. CXR-only models (CheXOne, CheXagent) are not evaluated on non-chest body parts.

\subsubsection{Open Question Datasets}
\label{sec:open}

The closed-format results above isolate discrete clinical decisions, but they constrain the model to a fixed set of options and so understate the harder problem a radiologist actually faces: producing free text that is both clinically correct and well formed. We therefore turn to open-ended evaluation, where each model generates unconstrained descriptions or full reports that are scored primarily with the Findings-Diagnosis methodology (Section~\ref{sec:evaluation}), alongside conventional report-generation metrics stated in the literature. As discussed earlier, surface-level text-overlap scores can diverge sharply from clinical correctness, so the Findings-Diagnosis score is treated as the primary signal and the remaining metrics as context.

The following subsections report open-ended performance across four complementary settings: clinician-authored cases spanning multiple body parts and question types (RadBench, Section~\ref{sec:radbench}); large-scale public chest-X-ray report generation (RexGradient, Section~\ref{sec:rexgradient}); the internal, multi-body-part held-out set (RadCoverage-VQA, Section~\ref{sec:radcoverage}); and mammography (Section~\ref{sec:mammo}). Together they trace HR1.5's open-ended behaviour from standardised, clinician-curated questions through to everyday reporting across the breadth of plain-film radiology.

\subsubsection*{RadBench Evaluation}
\label{sec:radbench}

RadBench is a clinician-authored evaluation framework developed at harrison.ai for benchmarking multimodal radiology foundation models. It consists of 105 unique X-ray cases with 497 questions (378 closed, 119 open) spanning question categories including finding presence, anatomical location, laterality, diagnosis, clinical reasoning, normal/abnormal classification, and modality identification.

\textbf{Closed-format evaluation} uses binary accuracy, precision, recall, F1, and Matthews correlation for yes/no questions, and accuracy with option-count-scaled weighting for multi-choice questions. \textbf{Open-ended evaluation} uses ROUGE \citep{Lin2004}, BLEU \citep{Papineni2002}, CheXBERT \citep{Smit2020}, RadGraph F1, the Findings-Diagnosis score described above, and an LLM-as-a-judge \citep{Gu2024} adjudication (Judge F-D).

\textbf{A note on data availability.} A portion of the RadBench cases is sourced from the NIH MedPix teaching atlas. Since RadBench was released, MedPix has been \emph{permanently retired} by the U.S.\ National Library of Medicine (NLM), following a realignment of agency priorities and resources; this is a permanent withdrawal rather than a temporary outage. As a consequence, many of the source images can no longer be retrieved, and the open-ended split has shrunk to only 69 usable cases. This remaining sample is too small to support an unbiased comparison across models. We therefore report the RadBench open-ended results below for completeness and posterity only, and caution against reading them as a reliable ranking; the larger-scale evaluations in the surrounding sections (ReXGradient, RadCoverage-VQA, and mammography), together with the FRCR simulations, provide the load-bearing evidence.

\begin{table}[H]
\centering
\caption{RadBench open-ended evaluation on the reduced (n=69) set. Reported for completeness only; see the note on data availability above.}
\label{tab:radbench}
\small
\begin{adjustbox}{max width=\textwidth}
\begin{tabular}{@{}lccccccc@{}}
\toprule
\textbf{Model} & \textbf{RadGraph F1 \up} & \textbf{RadCliQ-v1 \dn} & \textbf{F-D Score \up} & \textbf{CheXBERT \up} & \textbf{BLEU \up} & \textbf{ROUGE-L \up} & \textbf{Judge F-D \up} \\
\midrule
HR1.5+ & 0.008 & 0.274 & 58.0\% & 0.793 & 0.017 & 0.201 & 31.0\% \\
HR1.5 & \textbf{0.020} & 0.133 & 73.9\% & 0.858 & 0.049 & 0.470 & \textbf{37.1\%} \\
HR1 & 0.003 & 0.287 & 59.4\% & 0.788 & 0.000 & 0.173 & 28.4\% \\
\midrule
GPT-5.4 & 0.006 & 0.112 & \textbf{78.3\%} & 0.885 & 0.053 & 0.497 & 33.3\% \\
Gemini-3-Flash-Preview & 0.007 & \textbf{0.018} & 71.0\% & \textbf{0.921} & \textbf{0.085} & \textbf{0.608} & 35.1\% \\
Claude Opus 4.7 & 0.006 & 0.460 & 31.9\% & 0.654 & 0.002 & 0.065 & 24.9\% \\
MedGemma-27B & 0.000 & 0.149 & 59.4\% & 0.872 & 0.036 & 0.506 & 12.8\% \\
MedGemma-4B & 0.000 & 0.122 & 58.0\% & 0.886 & 0.000 & 0.489 & 16.5\% \\
\bottomrule
\end{tabular}
\end{adjustbox}
\end{table}

Table~\ref{tab:radbench} shows open-ended evaluation results using RadBench. Gemini-3-Flash-Preview achieves the best scores on most surface metrics, including RadCliQ-v1 (0.018), CheXBERT (0.921), BLEU (0.085), and ROUGE-L (0.608), while GPT-5.4 leads on the F-D Score (78.3\%). HR1.5 attains the highest Judge F-D score (37.1\%) and the highest RadGraph F1 (0.020). Claude Opus 4.7 records the lowest F-D Score (31.9\%), possibly because its responses diverge from the expected radiology format. Given the reduced sample (n=69; see the note above), these results are reported for completeness only.

A recurring cause of the divergence between surface metrics and clinical quality is easy to see at the level of individual items: a longer, more specific response can reduce exact-match overlap even as it improves clinical fidelity. To illustrate, consider the question ``What complication of the oesophageal stent has occurred?'', for which the ground truth is ``the oesophageal stent has migrated.'' HR1 answers: ``Complication of the oesophageal stent is evident. The oesophageal stent appears migrated.'' HR1.5 answers: ``The complication of the oesophageal stent is that it has slipped out of its original position and now lies horizontally in the fundus of the stomach.''

Under F-D scoring, HR1 is marked correct and HR1.5 incorrect. Yet HR1.5 is the clinically richer response: it not only identifies the migration but localizes the displaced stent as lying horizontally in the gastric fundus. The penalty arises purely from surface form. HR1.5 paraphrases ``migrated'' as ``slipped out of its original position,'' failing to reproduce the ground-truth keyword, whereas HR1's shorter answer echoes it verbatim. The longer, more specific output is thus scored as wrong despite its higher clinical fidelity.

\subsubsection*{RexGradient}
\label{sec:rexgradient}

RexGradient \citep{Zhang2025} is a large-scale chest X-ray evaluation benchmark comprising approximately 160,000 studies. We evaluate on the Zero-shot report generation task i.e.\ given patient demographics and clinical indication, generate a detailed findings statement.

Similar to the RadBench evaluation, performance is measured using RadGraph F1, CheXBERT \citep{Smit2020}, RadCliQ \citep{Yu2023} composite scores, and our Findings-Diagnosis methodology. Additionally, we evaluate over 11,000 closed-format questions derived from the dataset, spanning diagnosis (83\%), finding presence, normal/abnormal, clinical diagnosis, report interpretation, and laterality categories.

\begin{table}[H]
\centering
\caption{RexGradient zero-shot report generation.}
\label{tab:rexgradient}
\footnotesize
\begin{adjustbox}{max width=\textwidth}
\begin{tabular}{@{}lccccccc@{}}
\toprule
\textbf{Model} & \textbf{F-D Score \up} & \textbf{Judge F-D \up} & \textbf{RadGraph F1 \up} & \textbf{RadCliQ-v1 \dn} & \textbf{CheXBERT \up} & \textbf{BLEU \up} & \textbf{ROUGE-L \up} \\
\midrule
HR1.5 & \makecell{49.7\%\\(48.8--50.7)} & \makecell{68.9\%\\(68.0--69.8)} & 0.124 & 0.407 & 0.450 & 0.042 & 0.204 \\
HR1.5+ & \makecell{43.1\%\\(42.2--44.1)} & \makecell{\textbf{73.0\%}\\\textbf{(71.7--73.5)}} & 0.142 & 0.378 & 0.471 & 0.039 & 0.179 \\
HR1 & \makecell{46.7\%\\(45.7--47.7)} & \makecell{63.5\%\\(62.6--64.5)} & 0.149 & 0.384 & 0.443 & 0.044 & 0.199 \\
CheXOne-Reasoning & \makecell{47.1\%\\(46.1--48.0)} & \makecell{65.0\%\\(64.1--66.0)} & \textbf{0.238} & \textbf{0.158} & 0.529 & \textbf{0.110} & \textbf{0.281} \\
CheXOne-Instruct & \makecell{47.7\%\\(46.7--48.7)} & \makecell{66.6\%\\(65.7--67.5)} & 0.232 & 0.167 & \textbf{0.530} & 0.108 & 0.277 \\
MAIRA-2 & \makecell{46.7\%\\(45.7--47.7)} & \makecell{65.3\%\\(64.4--66.2)} & 0.152 & 0.329 & 0.491 & 0.073 & 0.224 \\
MedGemma-27B & \makecell{47.0\%\\(46.0--48.0)} & \makecell{60.6\%\\(59.7--61.6)} & 0.166 & 0.342 & 0.460 & 0.068 & 0.243 \\
MedGemma-4B & \makecell{\textbf{50.5\%}\\\textbf{(49.6--51.5)}} & \makecell{61.0\%\\(60.0--61.9)} & 0.208 & 0.266 & 0.476 & 0.070 & 0.243 \\
CheXagent & \makecell{25.7\%\\(24.9--26.6)} & \makecell{42.0\%\\(41.1--43.0)} & 0.055 & 0.650 & 0.336 & 0.020 & 0.126 \\
\bottomrule
\end{tabular}
\end{adjustbox}
\end{table}

Table~\ref{tab:rexgradient} shows zero-shot report generation results. RadCliQ-v1 is a composite error metric (lower is better). CheXOne-Reasoning achieves the highest RadGraph F1 (0.238), RadCliQ-v1 (0.158), BLEU, and ROUGE-L, while CheXOne-Instruct edges ahead on CheXBERT (0.530). MedGemma-4B leads on the automated Findings-Diagnosis score (50.5\%), while on the Judge Findings-Diagnosis score --- the more reliable measure on this benchmark --- HR1.5+ leads (73.0\%). The gap between text-overlap metrics and F-D score highlights the distinction between surface-level textual similarity and clinical accuracy.

A note on this benchmark: a substantial portion of RexGradient consists of structured reports with additional sections and finer-grained nodes than a free-text findings statement. On such references the automated Findings-Diagnosis score performs suboptimally due to limitations of its underlying RadGraph entity extraction, whose finding-matching becomes brittle when the ground truth is heavily structured (for the reasons set out in Section~\ref{sec:evaluation}), and the F-D column above reflects this. The Judge Findings-Diagnosis score is more representative of clinical correctness here, with HR1.5+ attaining the highest value (73.0\%); even so, a residual gap remains under this stronger evaluator, underscoring that no current automatic evaluator fully captures report quality on structured data.

\subsubsection*{RadCoverage-VQA}
\label{sec:radcoverage}

RadCoverage-VQA is an internal, held-out evaluation set of clinical questions built for broad body-part coverage (chest X-ray, MSK, spine, abdomen, pelvis, and mammography) across a mixed set of medium-to-hard clinical tasks. Cases are drawn from validation-tagged production data, providing a diverse and clinically representative test set that extends well beyond chest-only evaluation.

Performance is measured using the Findings-Diagnosis methodology, with results broken down by body-part category. Body parts are grouped using the same mapping as the closed datasets (Section~\ref{sec:closed}): CXR, MSK, Abdomen, and Breast, with remaining body parts (clavicle, coccyx, fibula, finger, sacrum) in Other.

\begin{table}[H]
\centering
\caption{RadCoverage-VQA Findings-Diagnosis scores by body-part group.}
\label{tab:radcoverage}
\footnotesize
\begin{adjustbox}{max width=\textwidth}
\begin{tabular}{@{}lcccccc@{}}
\toprule
\textbf{Model} & \makecell{\textbf{CXR}\\\textbf{(n=3224)}} & \makecell{\textbf{MSK}\\\textbf{(n=10245)}} & \makecell{\textbf{Abdomen}\\\textbf{(n=1805)}} & \makecell{\textbf{Breast}\\\textbf{(n=944)}} & \makecell{\textbf{Other}\\\textbf{(n=340)}} & \makecell{\textbf{Overall}\\\textbf{(n=16558)}} \\
\midrule
HR1.5 & \makecell{\textbf{70.1\%}\\\textbf{(68.5--71.7)}} & \makecell{\textbf{64.8\%}\\\textbf{(63.9--65.7)}} & \makecell{\textbf{70.0\%}\\\textbf{(67.8--72.1)}} & \makecell{\textbf{67.7\%}\\\textbf{(64.7--70.6)}} & \makecell{\textbf{65.3\%}\\\textbf{(60.1--70.2)}} & \makecell{\textbf{66.6\%}\\\textbf{(65.9--67.3)}} \\
HR1.5+ & \makecell{50.2\%\\(48.5--51.9)} & \makecell{47.6\%\\(46.6--48.6)} & \makecell{36.5\%\\(34.3--38.7)} & \makecell{43.5\%\\(40.4--46.7)} & \makecell{47.9\%\\(42.6--53.2)} & \makecell{46.6\%\\(45.8--47.4)} \\
HR1 & \makecell{65.9\%\\(64.2--67.5)} & \makecell{63.1\%\\(62.2--64.0)} & \makecell{66.9\%\\(64.7--69.0)} & \makecell{67.5\%\\(64.4--70.4)} & \makecell{60.0\%\\(54.7--65.1)} & \makecell{64.3\%\\(63.6--65.0)} \\
MedGemma-27B & \makecell{17.4\%\\(16.1--18.7)} & \makecell{15.7\%\\(15.0--16.4)} & \makecell{10.7\%\\(9.4--12.2)} & \makecell{13.1\%\\(11.1--15.4)} & \makecell{14.1\%\\(10.8--18.2)} & \makecell{15.3\%\\(14.8--15.9)} \\
MedGemma-4B & \makecell{25.5\%\\(24.0--27.0)} & \makecell{22.9\%\\(22.1--23.7)} & \makecell{16.0\%\\(14.4--17.8)} & \makecell{16.5\%\\(14.3--19.0)} & \makecell{26.5\%\\(22.1--31.4)} & \makecell{22.4\%\\(21.8--23.0)} \\
\bottomrule
\end{tabular}
\end{adjustbox}
\end{table}

Table~\ref{tab:radcoverage} shows Findings-Diagnosis scores on RadCoverage-VQA. HR1.5 leads across all body-part categories, with the largest gains over HR1 in Abdomen (+3.1pp) and CXR (+4.2pp). MedGemma models score substantially lower, likely because RadCoverage-VQA uses a multi-turn dialogue format that differs from their training distribution.

\subsubsection*{Mammography}
\label{sec:mammo}

Mammography evaluation extends the Findings-Diagnosis framework to breast imaging using the \href{https://www.cancerimagingarchive.net/collection/cbis-ddsm/}{CBIS-DDSM} dataset (n=1625).

\begin{table}[H]
\centering
\caption{Mammography (CBIS-DDSM) Findings-Diagnosis scores.}
\label{tab:mammo}
\begin{tabular}{@{}lcc@{}}
\toprule
\textbf{Model} & \textbf{F-D Score \up} & \textbf{95\% CI} \\
\midrule
\textbf{HR1.5+} & \textbf{77.8\%} & 75.6--79.7\% \\
HR1.5 & 64.9\% & 62.5--67.1\% \\
HR1 & 64.9\% & 62.6--67.2\% \\
\midrule
MedGemma-27B & 67.6\% & 65.3--69.9\% \\
MedGemma-4B & 61.4\% & 59.0--63.7\% \\
\bottomrule
\end{tabular}
\end{table}

Table~\ref{tab:mammo} shows mammography Findings-Diagnosis scores. HR1.5+ leads at 77.8\%, a substantial improvement over HR1.5 and HR1 (both 64.9\%). MedGemma-27B achieves 67.6\%, outperforming HR1.5 and HR1 on this modality.
\section{Qualitative Results}
\label{sec:qual}

Beyond aggregate benchmark scores, we also characterised qualitative aspects of HR1.5's output through structured radiologist review. This section organises that assessment around six capabilities that together characterise how HR1.5 performs as compared to publicly available frontier models.

We compare (dated May, 2026) HR1.5+ (Harrison.Rad~1.5, agent-driven VQA, temperature 0) against three publicly available frontier multimodal models: GPT-5.4 (OpenAI), Gemini-3-Flash-Preview (Google) and Claude Opus 4.7 (bedrock-anthropic-claude-opus-4-7, AWS Bedrock). Decoding temperature was set to 0 wherever the model accepts it (HR1.5+, GPT-5.4 and Gemini-3-Flash-Preview) and left at the provider default otherwise: Claude Opus 4.7 (which does not respect a temperature parameter). All remaining sampling parameters were left at provider defaults. Every model received the same prompt and image(s) per case.

\subsection{Accurate Findings Detection}

Accurate image interpretation is core to the function of a radiology foundation model: correctly identifying the finding that is actually present, especially when it is subtle and easily mistaken for a more dramatic abnormality.

HR1.5+ names the scaphoid waist fracture directly. All three frontier models miss it, each with a different wrong read --- GPT-5.4 a soft-tissue mass with ``bones intact,'' Gemini a giant-cell tumour of the tendon sheath, and Claude Opus 4.7 a distal radius (Colles') fracture (figure 3.1).

\begin{figure}[htbp]
  \centering
  \includegraphics[height=0.82\textheight,width=\linewidth,keepaspectratio]{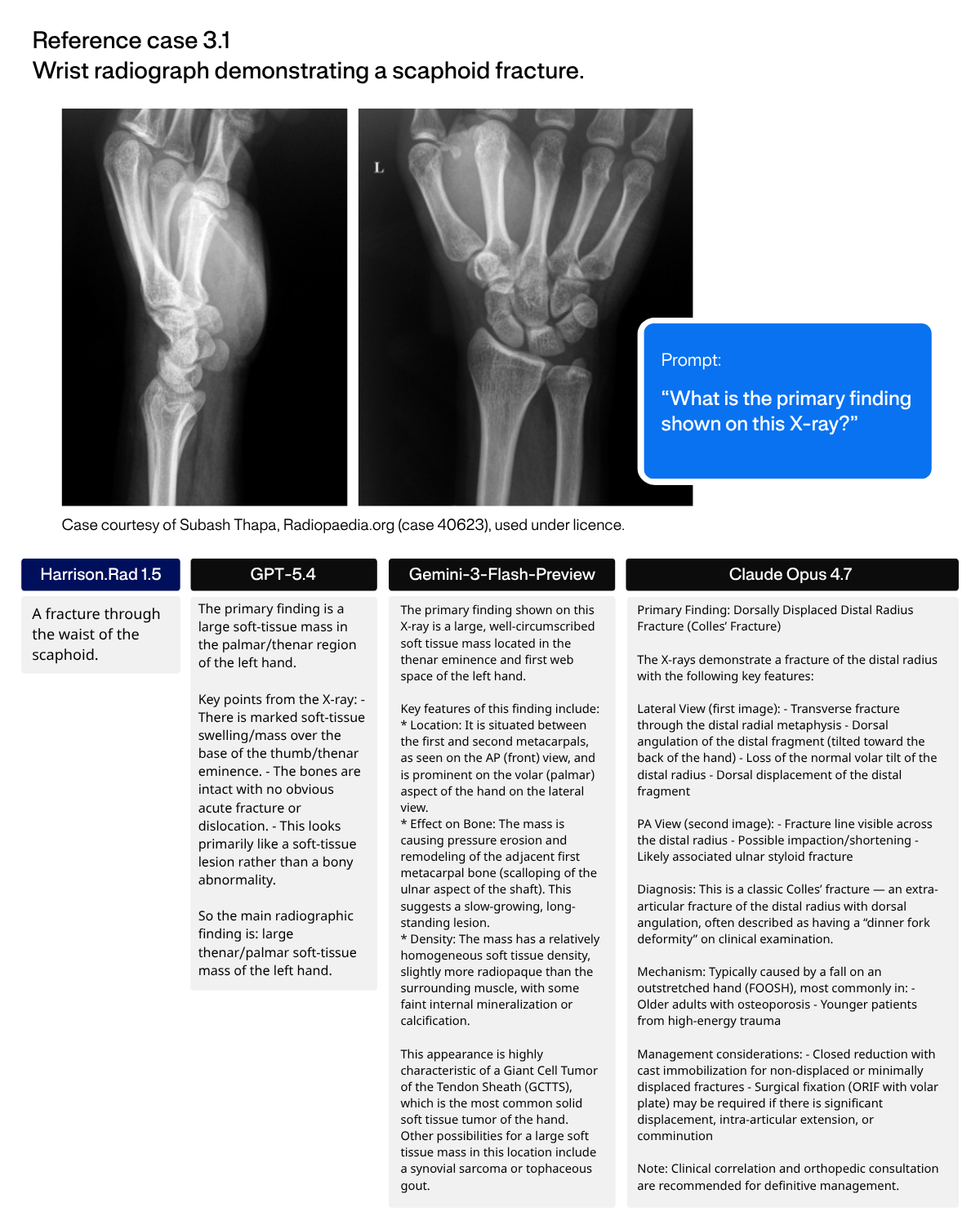}
  \caption*{\textbf{Figure 3.1.} Scaphoid fracture. Case courtesy of Subash Thapa, Radiopaedia.org (case \href{https://radiopaedia.org/cases/40623}{40623}), used under licence.}
\end{figure}

\subsection{Draft Report Generation}

The primary requirement is that the draft be accurate and faithfully to the findings, and correctness is never traded off against style. Given an accurate draft, the secondary aim is to present it in a form close to what the reporting radiologist would have written, so the content reads naturally and reduces cognitive load rather than expecting them to change practice. Crucially, matching the radiologist's prose is intended to ease review, not to substitute for it. The draft is a starting point that the radiologist verifies and edits, and familiarity of style should make discrepancies easier to spot rather than encouraging uncritical acceptance.

Reference case 3.2, a portable chest radiograph demonstrating innumerable bilateral pulmonary metastases with an indwelling PICC line. The verbatim prompt and each model's verbatim answer are shown below, first to the image alone (figure 3.2.1), then with the clinical history added (drafting a report is exactly where history should be used), with the key finding in each answer highlighted in bold (figure 3.2.2).

From the image alone, HR1.5+ names the metastases directly, while all three frontier models read the identical film as pneumonia / COVID-19.

\begin{figure}[htbp]
  \centering
  \includegraphics[height=0.82\textheight,width=\linewidth,keepaspectratio]{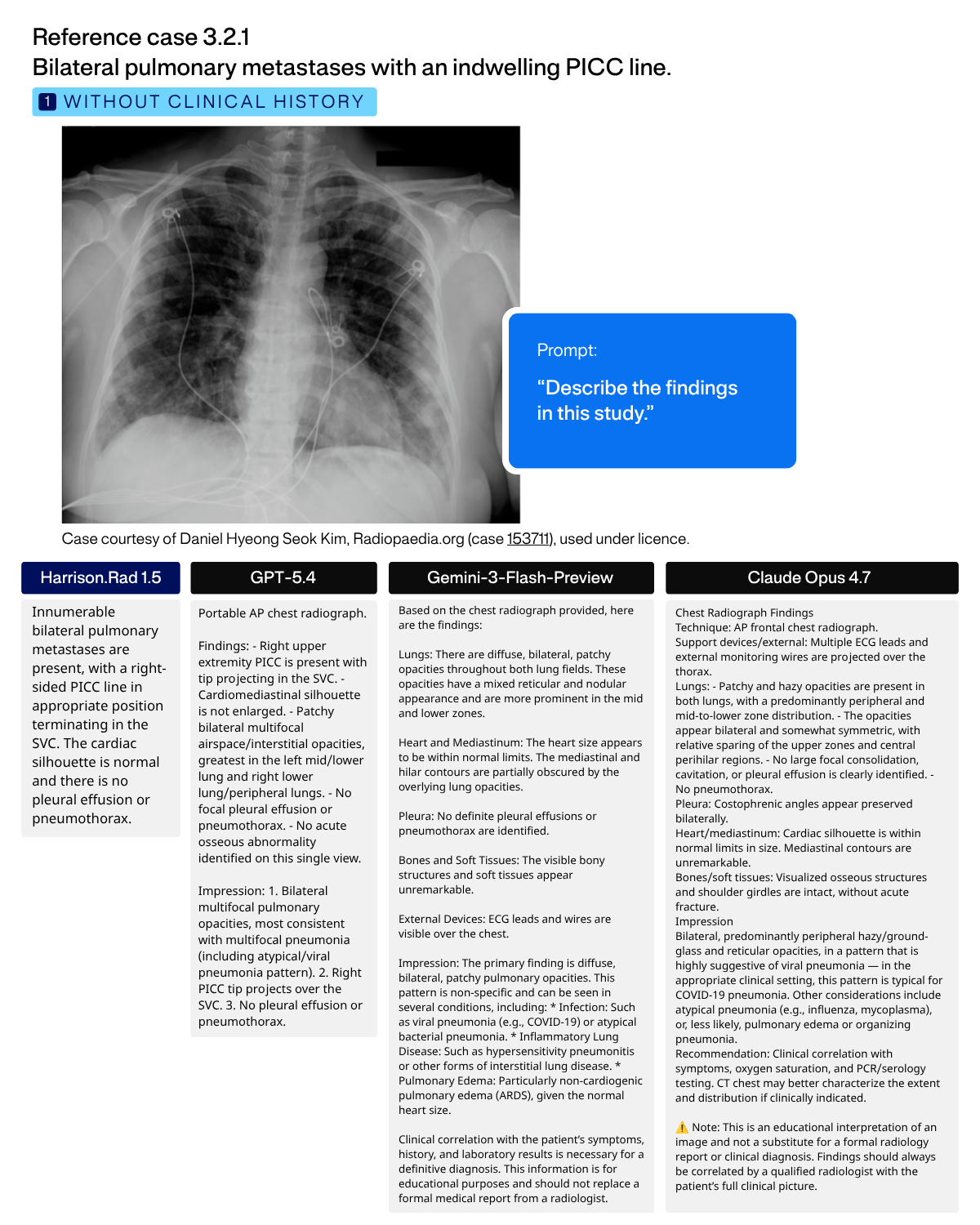}
  \caption*{\textbf{Figure 3.2.1.} Metastatic lung disease, portable chest radiograph. Case courtesy of Daniel Hyeong Seok Kim, Radiopaedia.org (case \href{https://radiopaedia.org/cases/153711}{153711}), used under licence. Image only (no clinical history).}
\end{figure}

\begin{figure}[htbp]
  \centering
  \includegraphics[height=0.82\textheight,width=\linewidth,keepaspectratio]{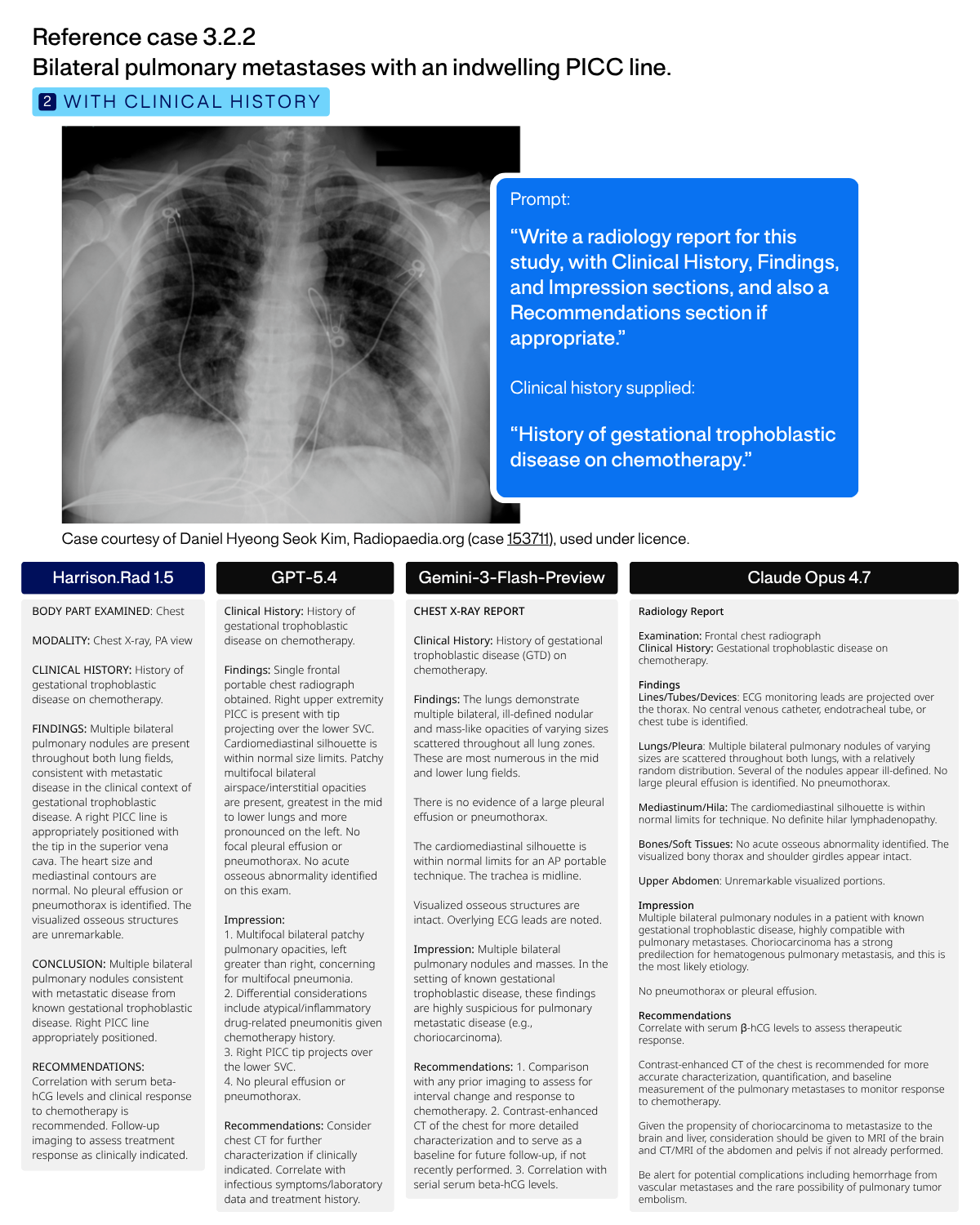}
  \caption*{\textbf{Figure 3.2.2.} Metastatic lung disease, portable chest radiograph. Case courtesy of Daniel Hyeong Seok Kim, Radiopaedia.org (case \href{https://radiopaedia.org/cases/153711}{153711}), used under licence. Image with clinical history.}
\end{figure}

With the history supplied, HR1.5+, Gemini-3-Flash-Preview and Claude Opus 4.7 all identify the bilateral pulmonary metastases, and HR1.5+ additionally reports the appropriately-sited PICC line. GPT-5.4 still reads the nodules as multifocal pneumonia and never mentions metastatic disease, a persistent miss even when handed the diagnosis-relevant history.

\subsection{Longitudinal Reasoning with Priors}

When prior imaging is available, HR1.5 automatically compares the current study against it and surfaces interval change, described in plain clinical language rather than as a flagged checklist. The comparison is performed at the moment the case is opened, so the report is contextualised against the patient's history from the outset.

Reference case 3.3, current and prior chest radiographs (supplied labelled), with the clinical history ``Day 2 of antibiotics, follow-up imaging.'' The reference optimal report describes interval worsening of right-sided airspace opacification with a large right pleural effusion and new left mid-zone opacity (figure 3.3).

HR1.5+ reports the interval progression correctly, with increased right-sided airspace opacification, a new left mid-zone opacity and the enlarging right effusion, matching the reference, as does GPT-5.4. Gemini-3-Flash-Preview adds a spurious cardiomegaly, and Claude Opus 4.7 misses the large effusion and fabricates bilateral central venous catheters that are not present. Overt interval progression is thus within reach for HR1.5+; subtler change, for example a fracture occult on the prior and newly visible, is harder for a single pass, which the production comparison route addresses by reading each study independently and diffing the two.

\begin{figure}[htbp]
  \centering
  \includegraphics[height=0.82\textheight,width=\linewidth,keepaspectratio]{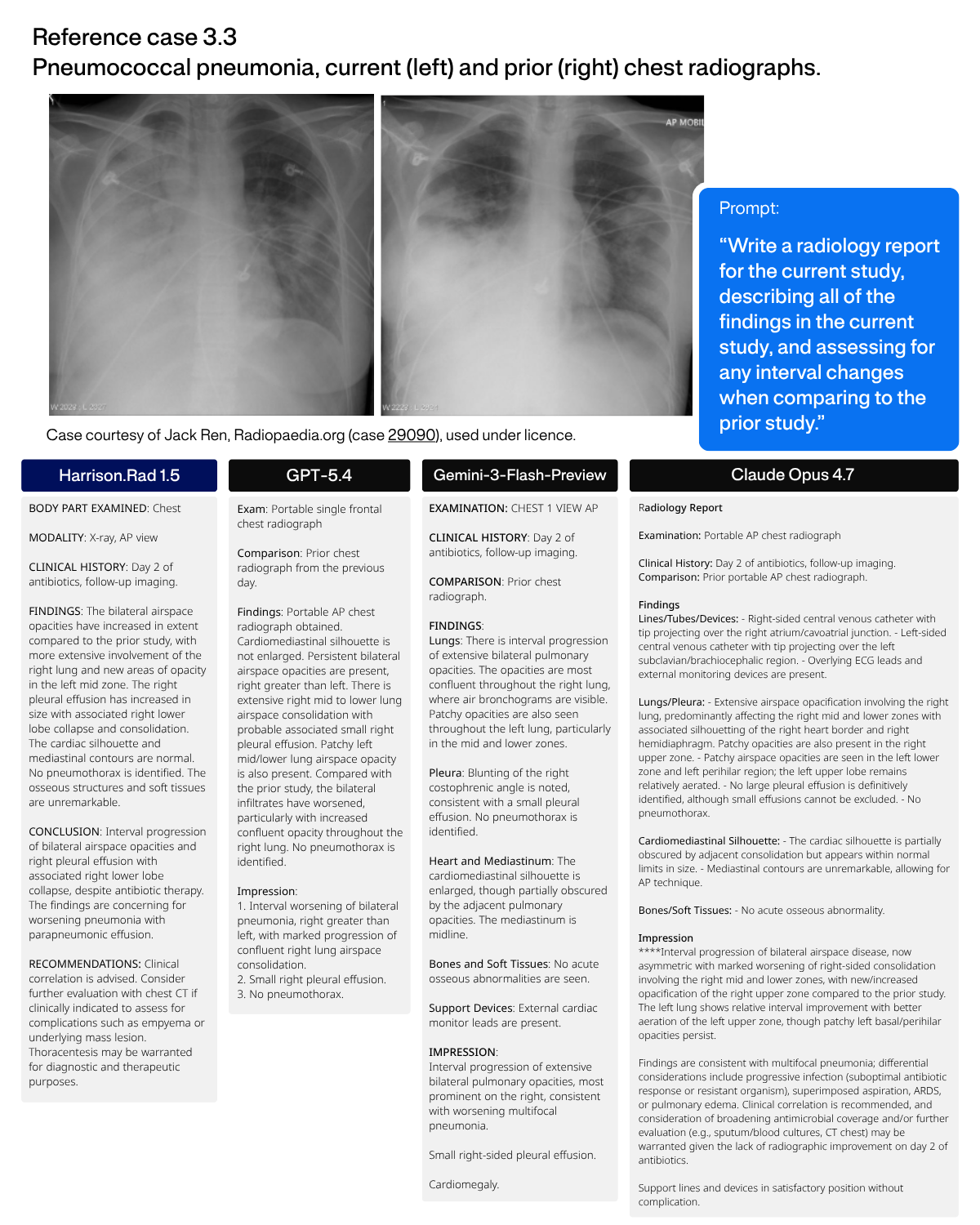}
  \caption*{\textbf{Figure 3.3.} Pneumococcal pneumonia, current (left) and prior (right) chest radiographs. Case courtesy of Jack Ren, Radiopaedia.org (case \href{https://radiopaedia.org/cases/29090}{29090}), used under licence.}
\end{figure}

\subsection{Precise Anatomical Localisation}

Findings are placed with greater anatomical specificity, including quantitative measurements such as the cardiothoracic ratio. The improvement is most pronounced on the spine, abdomen, and post-procedural imaging, among the areas of highest diagnostic complexity.

Reference case 3.4, a chest radiograph demonstrating a left apical mass in the left upper lobe, partially obscured by the clavicle (figure 3.4). HR1.5+ localises the mass correctly to the left upper lobe, as does Gemini-3-Flash-Preview; GPT-5.4 gets the side but mischaracterises it as a calcified mediastinal/chest-wall lesion. Claude Opus 4.7 lateralises it to the wrong (right) side, a laterality error with direct clinical consequences.

\begin{figure}[htbp]
  \centering
  \includegraphics[height=0.82\textheight,width=\linewidth,keepaspectratio]{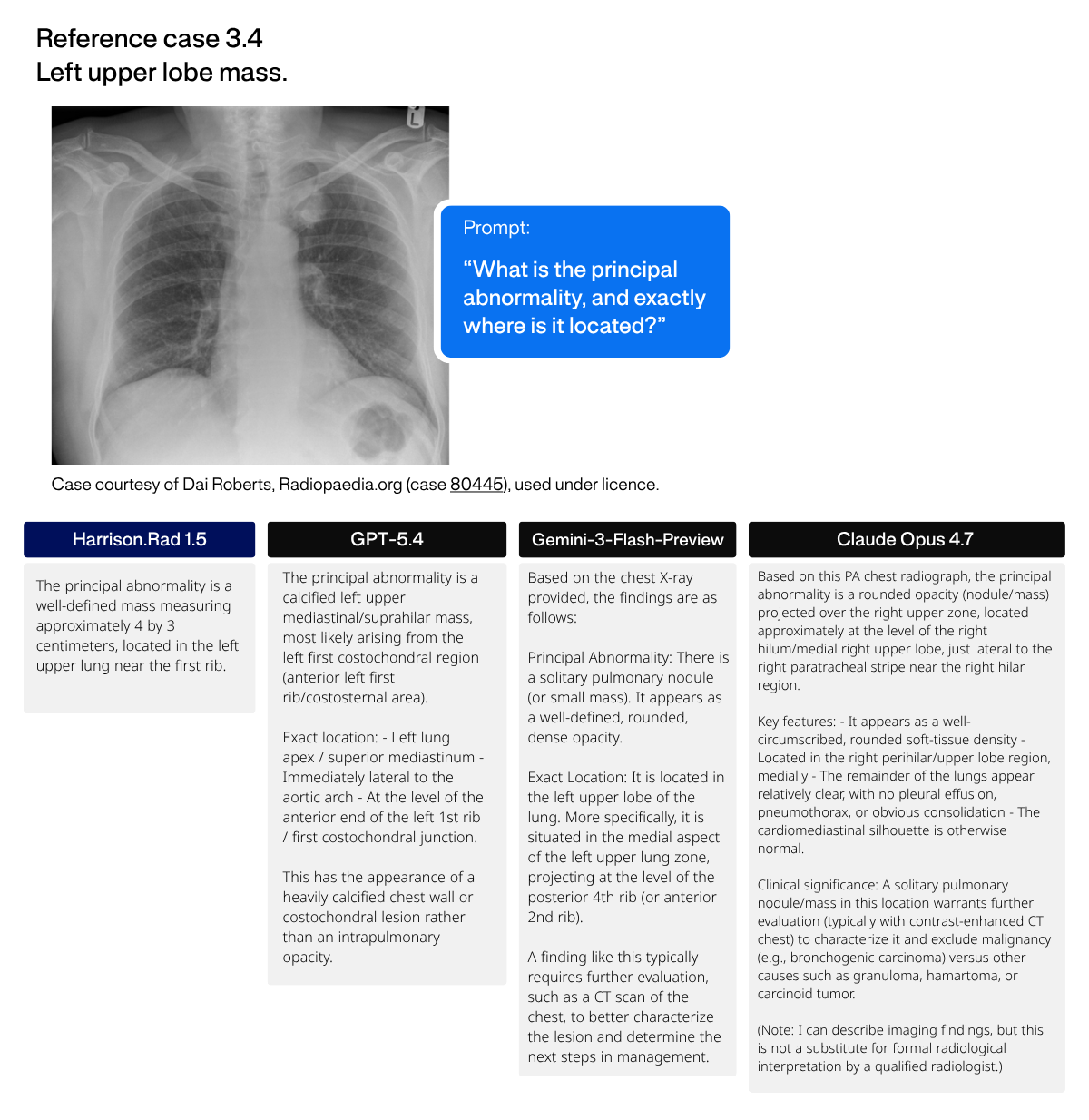}
  \caption*{\textbf{Figure 3.4.} Left upper lobe mass. Case courtesy of Dai Roberts, Radiopaedia.org (case \href{https://radiopaedia.org/cases/80445}{80445}), used under licence.}
\end{figure}

\subsection{Complex Multi-Finding Reasoning}

HR1.5 shows improved performance on high-cognitive-load studies that contain multiple findings, precisely the cases in which a finding is most likely to be missed. It also offers stronger coverage of rare-but-real findings that fall outside the classical reporting ontology.

Multiple coexisting findings (a chest study). Reference case 3.5.1, PA and lateral chest radiographs demonstrating congestive cardiac failure, a study whose several coexisting findings (cardiomegaly, pulmonary venous congestion, interstitial oedema) must be tied to a single diagnosis (figure 3.5.1).

HR1.5+ ties the cardiomegaly, pulmonary venous congestion and interstitial oedema together as congestive cardiac failure, as does GPT-5.4. Gemini-3-Flash-Preview instead diagnoses sarcoidosis, hallucinating bilateral hilar and paratracheal lymphadenopathy; Claude Opus 4.7 diagnoses combined pulmonary fibrosis and emphysema, calling the cardiomediastinal silhouette normal and so missing the cardiomegaly, two confident non-cardiac misreadings of a classic heart-failure film.

A complex single-region injury. Reference case 3.5.2, a knee radiograph series demonstrating three coexisting findings: a lateral patellar dislocation, a medial patellar avulsion fracture, and a joint effusion (figure 3.5.2).

HR1.5+ enumerates all three coexisting findings, as does Claude Opus 4.7 (describing the medial avulsion as an osteochondral fragment). GPT-5.4 mis-reads the displaced patella as a normal bipartite-patella variant, concluding ``no acute fracture or dislocation,'' while Gemini under-grades the dislocation as subluxation, the high-cognitive-load setting where a missed finding is most consequential.

\begin{figure}[htbp]
  \centering
  \includegraphics[height=0.82\textheight,width=\linewidth,keepaspectratio]{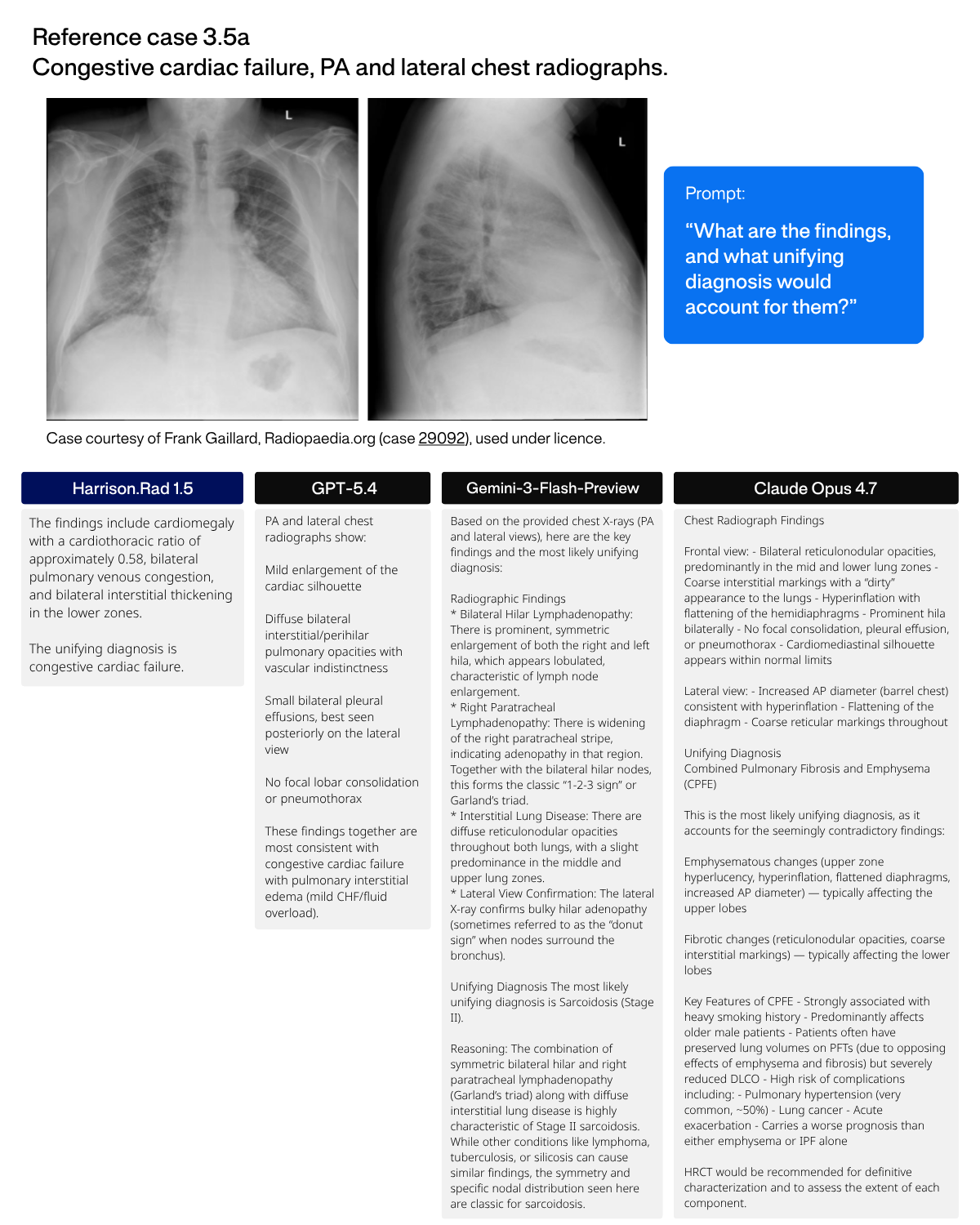}
  \caption*{\textbf{Figure 3.5.1.} Congestive cardiac failure, PA and lateral chest radiographs. Case courtesy of Frank Gaillard, Radiopaedia.org (case \href{https://radiopaedia.org/cases/29092}{29092}), used under licence.}
\end{figure}

\begin{figure}[htbp]
  \centering
  \includegraphics[height=0.82\textheight,width=\linewidth,keepaspectratio]{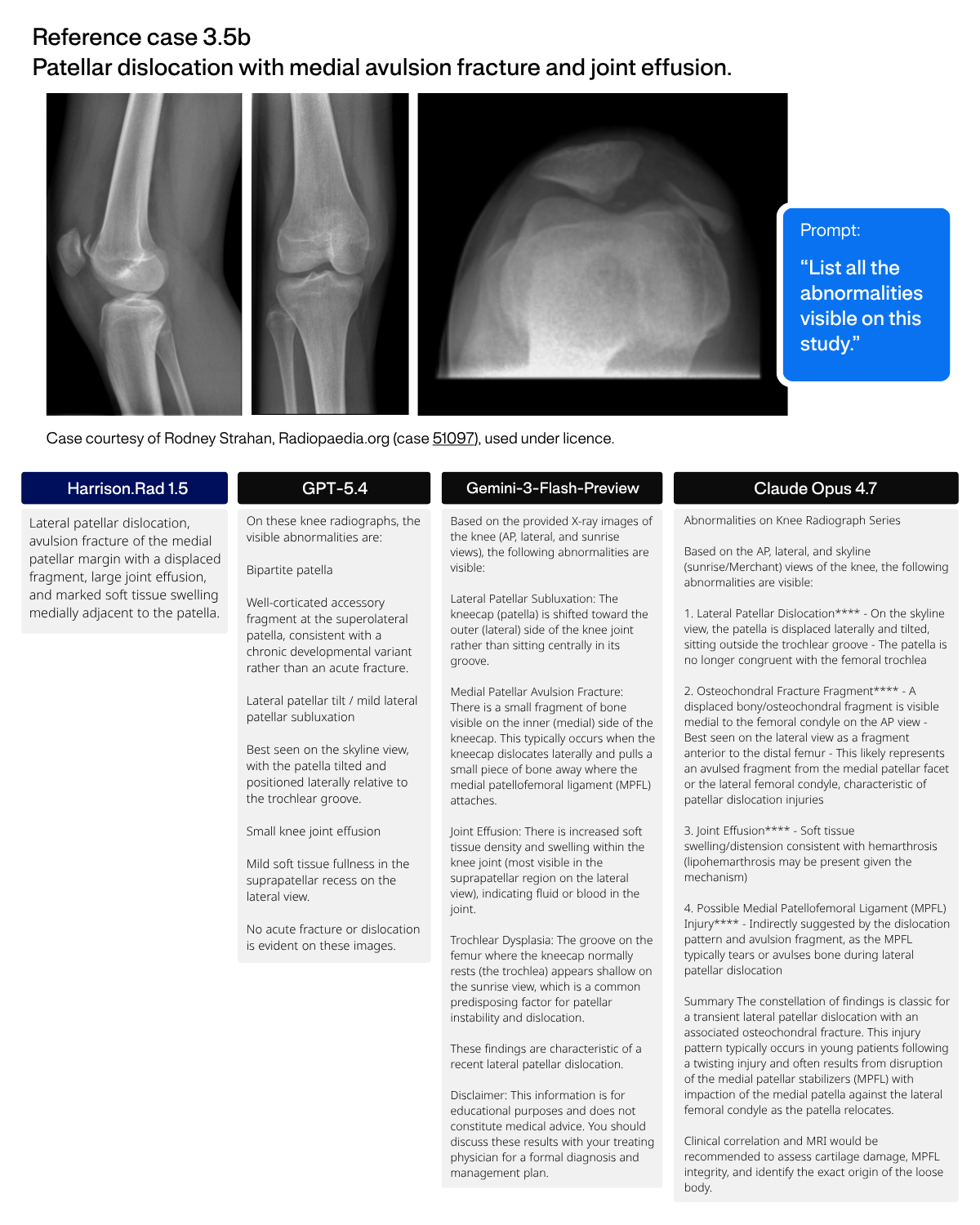}
  \caption*{\textbf{Figure 3.5.2.} Patellar dislocation with medial avulsion fracture and joint effusion. Case courtesy of Rodney Strahan, Radiopaedia.org (case \href{https://radiopaedia.org/cases/51097}{51097}), used under licence.}
\end{figure}

\FloatBarrier

\subsection{Style and Template Adaptation}

Slotting into an existing reporting workflow means following a prescribed report template exactly: populating the sections that apply and, just as importantly, suppressing those that do not, rather than padding the report with empty boilerplate. The test is therefore not whether a model can produce section headings (all of them can) but whether it treats the template's conditional instructions as rules to obey.

Reference case 3.6, as shown in figure 3.6, a chest radiograph demonstrating interstitial thickening (a fibrosis pattern), with the clinical history ``Shortness of breath'' supplied. The study is reported against a fixed sectioned template (mirroring the structured variant the rad-agent production pipeline emits) in which Lines \& Devices, Abdominal and Soft Tissue are explicitly marked optional, to be included only when relevant, while the remaining regions take a normal statement where no positive finding is present.

All four models detect the bilateral interstitial/fibrotic process, follow the capitalised heading convention, and produce the sectioned layout, so neither detection nor formatting separates them. The differentiator is the conditional instruction to include the optional sub-regions only when relevant. HR1.5+ and GPT-5.4 follow it exactly here, emitting only the populated regions and dropping all three optional sub-regions (Lines \& Devices, Abdominal, Soft Tissue), none of which apply. Claude Opus 4.7 appends a redundant Soft Tissue: Unremarkable, and Gemini-3-Flash-Preview appends both Abdominal and Soft Tissue.

\begin{figure}[htbp]
  \centering
  \includegraphics[height=0.82\textheight,width=\linewidth,keepaspectratio]{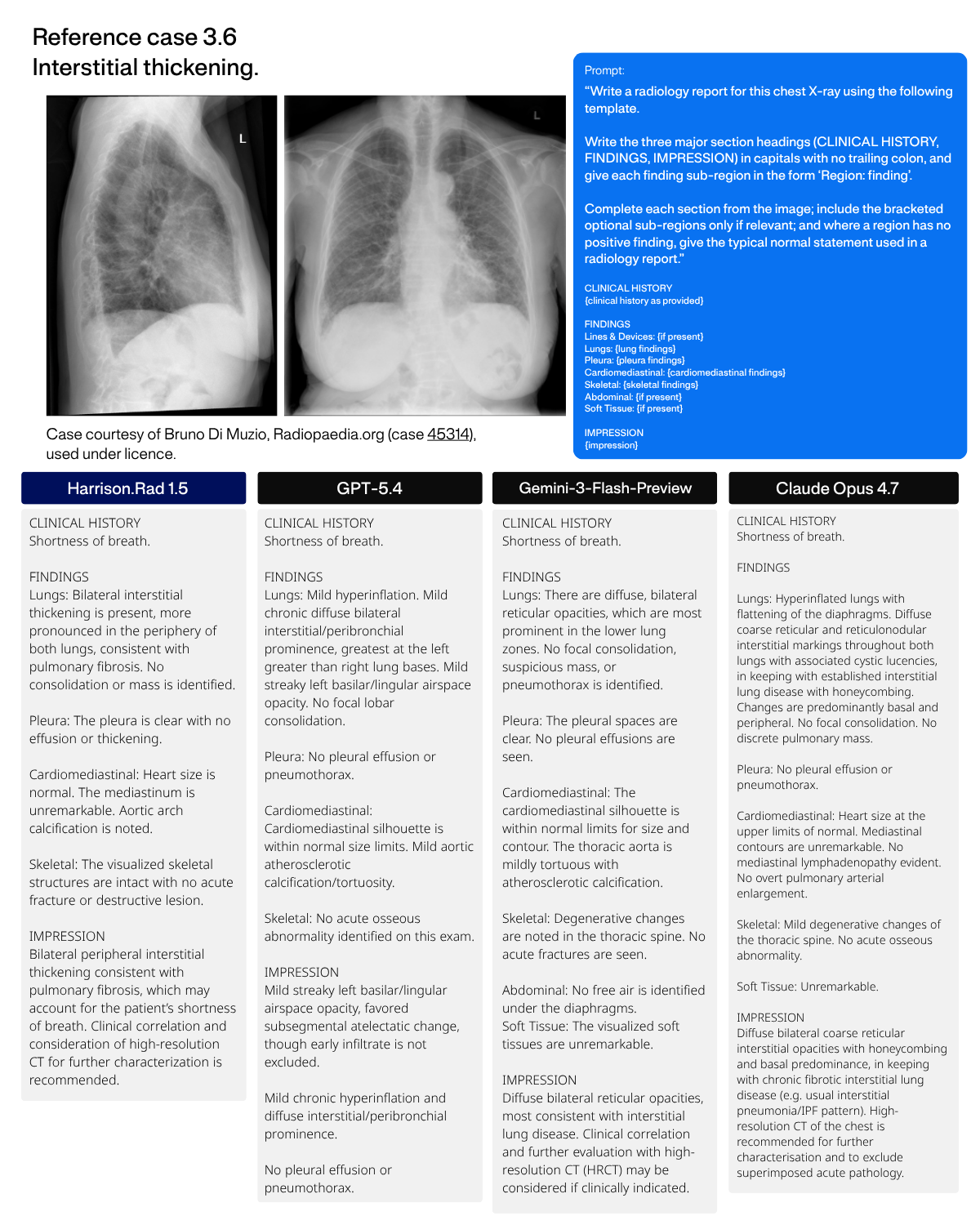}
  \caption*{\textbf{Figure 3.6.} Interstitial thickening. Case courtesy of Bruno Di Muzio, Radiopaedia.org (case \href{https://radiopaedia.org/cases/45314}{45314}), used under licence.}
\end{figure}

\FloatBarrier
\section{Explainability}
\label{sec:explain}

Instruction-tuned vision-language models tend to present their answers as fluent text without surfacing the basis for them, which makes them harder to interpret. We see value in complementing HR1.5's outputs with explanations that show the reasoning behind an answer and connect it back to verifiable features in the image. Such grounding makes the model easier to adopt and to work with \citep{Reyes2020,Tonekaboni2019}: it lets a reader trace why an answer was given, form a view of how reliable it is \citep{Hendricks2016,Selvaraju2017}, and develop a clearer sense of where the model's strengths and limits lie \citep{Ghassemi2021}. The methods described below are exploratory, early steps in this direction for HR1.5 rather than a finished or validated capability, and they treat the model's own internal signals as additional evidence to be surfaced rather than discarded.

To this end, with HR1.5 we have built a post-hoc explainability engine that derives interpretable outputs alongside each prediction, using only quantities the model already computes during a normal forward pass (activations, gradients, and attention weights) with no retraining or architectural change. The engine is organised around three questions a clinician asks of any model output:

\begin{itemize}[leftmargin=1.4em]
  \item \textbf{Concept understanding} (Section~\ref{sec:concept}): \textit{Does the model truly know this finding?} We test whether findings are genuinely encoded and used internally, not just named.
  \item \textbf{Concept localisation} (Section~\ref{sec:localisation}): \textit{Where on the image is this answer coming from?} We map each answer back to the image regions that drove it.
  \item \textbf{Confidence and certainty} (Section~\ref{sec:confidence}): \textit{How much should I trust this answer?} We attach a reliability score to each answer and correct it for known sources of false certainty.
\end{itemize}

\subsection{Concept Understanding}
\label{sec:concept}

A central question for any clinical model is whether it genuinely represents the findings it reports. A linear probe that separates positive from negative examples shows that a concept is decodable from the model's activations, but decoding alone does not establish that the model uses that concept when forming an answer. The distinction matters clinically: only a finding the model actively uses, rather than one we can merely read off its activations, is a reliable handle for monitoring or intervention. To examine this we apply an offline analysis we term concept tracing, which combines four lines of evidence: extraction of a concept direction, validation of its separability, characterisation of its geometry and depth, and a causal steering test. The results below cover the two largest free-text radiography domains, chest radiography (CXR) and musculoskeletal imaging (MSK).

For each domain we curate a vocabulary of clinically salient findings and extract, for each, a unit-norm linear direction from labelled activations. The primary estimator is a difference of means, the unit-normalised difference between the mean positive and mean negative activations, which has been shown to be more causally implicated in model outputs than more elaborate probes \citep{Marks2023}; L2-regularised logistic regression \citep{Alain2017} serves as an alternative. Two extraction sites are probed independently: the decoder residual stream at the last prompt token, where the model has integrated the full image and question, and the image-side latent representation pooled across the bridger's slots, which indicates whether a concept is already linear before text conditioning. Probes are trained within each domain so that a direction discriminates the finding itself rather than the imaging modality.

On a held-out split, each concept is scored by its positive-to-negative projection gap, ROC-AUC, and Cohen's $d$, and assigned to the decoder layer at which its probe is strongest. Across the full vocabularies, both domains separate held-out data well: 17 CXR concepts reach a mean ROC-AUC of 0.873 (median 0.853, mean Cohen's $d$ 1.82) and 24 MSK concepts reach 0.895 (median 0.907, mean Cohen's $d$ 1.91). Tables~\ref{tab:cxr-sep} and~\ref{tab:msk-sep} report every concept tested in each domain, with its held-out ROC-AUC and Cohen's $d$.

\begin{table}[htbp]
\centering
\begin{minipage}[t]{0.48\linewidth}
\centering
\caption{Held-out concept separability for CXR (ROC-AUC and Cohen's $d$).}
\label{tab:cxr-sep}
\small
\begin{adjustbox}{max width=\linewidth}
\begin{tabular}{@{}lcc@{}}
\toprule
\textbf{Concept} & \textbf{ROC-AUC} & \textbf{Cohen's $d$} \\
\midrule
scoliosis & 0.999 & 3.26 \\
hyperinflation & 0.995 & 3.43 \\
endotracheal tube & 0.965 & 2.30 \\
pacemaker & 0.956 & 2.10 \\
rib fracture & 0.934 & 3.16 \\
pleural thickening & 0.917 & 1.87 \\
subcutaneous emphysema & 0.915 & 2.52 \\
central venous catheter & 0.897 & 1.54 \\
pulmonary oedema & 0.853 & 1.46 \\
pneumonia & 0.843 & 1.55 \\
cardiomegaly & 0.832 & 1.27 \\
lung collapse & 0.799 & 1.06 \\
pneumothorax & 0.797 & 1.20 \\
ground glass opacity & 0.791 & 1.04 \\
atelectasis & 0.785 & 1.08 \\
consolidation & 0.781 & 1.05 \\
pleural effusion & 0.776 & 1.08 \\
\bottomrule
\end{tabular}
\end{adjustbox}
\end{minipage}\hfill
\begin{minipage}[t]{0.48\linewidth}
\centering
\caption{Held-out concept separability for MSK (ROC-AUC and Cohen's $d$).}
\label{tab:msk-sep}
\small
\begin{adjustbox}{max width=\linewidth}
\begin{tabular}{@{}lcc@{}}
\toprule
\textbf{Concept} & \textbf{ROC-AUC} & \textbf{Cohen's $d$} \\
\midrule
loose body & 0.999 & 3.22 \\
bone cyst & 0.985 & 3.32 \\
osteomyelitis & 0.969 & 2.81 \\
spiral fracture & 0.961 & 2.63 \\
comminuted fracture & 0.951 & 2.15 \\
joint replacement & 0.942 & 2.55 \\
displaced fracture & 0.916 & 1.74 \\
ankylosis & 0.915 & 1.62 \\
joint space narrowing & 0.914 & 1.80 \\
scoliosis & 0.913 & 1.96 \\
avulsion fracture & 0.912 & 1.98 \\
spondylosis & 0.907 & 1.66 \\
soft tissue swelling & 0.907 & 1.90 \\
spondylolisthesis & 0.903 & 1.66 \\
osteopenia & 0.883 & 1.64 \\
lytic lesion & 0.881 & 2.51 \\
kyphosis & 0.858 & 1.47 \\
degenerative change & 0.846 & 1.22 \\
sclerotic lesion & 0.845 & 1.18 \\
osteoarthritis & 0.842 & 1.49 \\
subluxation & 0.825 & 1.33 \\
dislocation & 0.813 & 1.27 \\
joint effusion & 0.803 & 1.53 \\
fracture & 0.800 & 1.22 \\
\bottomrule
\end{tabular}
\end{adjustbox}
\end{minipage}
\end{table}

These concepts span a clinically meaningful range, and the model encodes each as a linear direction that separates held-out data. Discrete, well-circumscribed findings are recovered with particular clarity: structural and device findings such as scoliosis, hyperinflation, the endotracheal tube, and the loose body and bone cyst in MSK exceed 0.96 with large effect sizes, and specific fracture morphologies (comminuted, spiral, displaced) and focal lesions (lytic lesion, osteomyelitis) are similarly well separated. Distributed, interpretation-heavy findings are encoded cleanly as well: pulmonary oedema, pneumonia, and cardiomegaly in CXR, and degenerative changes such as osteoarthritis, spondylosis, and joint space narrowing in MSK, all decode with ROC-AUC in the high 0.8s to low 0.9s despite being defined by diffuse and frequently co-occurring appearances, indicating that these directions capture the finding itself rather than a single localised cue.

Before asking whether the model uses these directions, it is worth examining how the directions relate to one another. We measure the pairwise cosine similarity between every pair of concept directions in each domain. The directions are distinct but structured: the mean absolute cosine similarity is 0.335 for CXR and 0.305 for MSK, and the first three principal components capture 59\% and 56\% of the variance respectively. The findings occupy separate axes rather than collapsing onto one, while still sharing a compact, low-dimensional geometry (Figures~4.1 and~4.2).

\begin{figure}[htbp]
  \centering
  \includegraphics[width=0.8\linewidth]{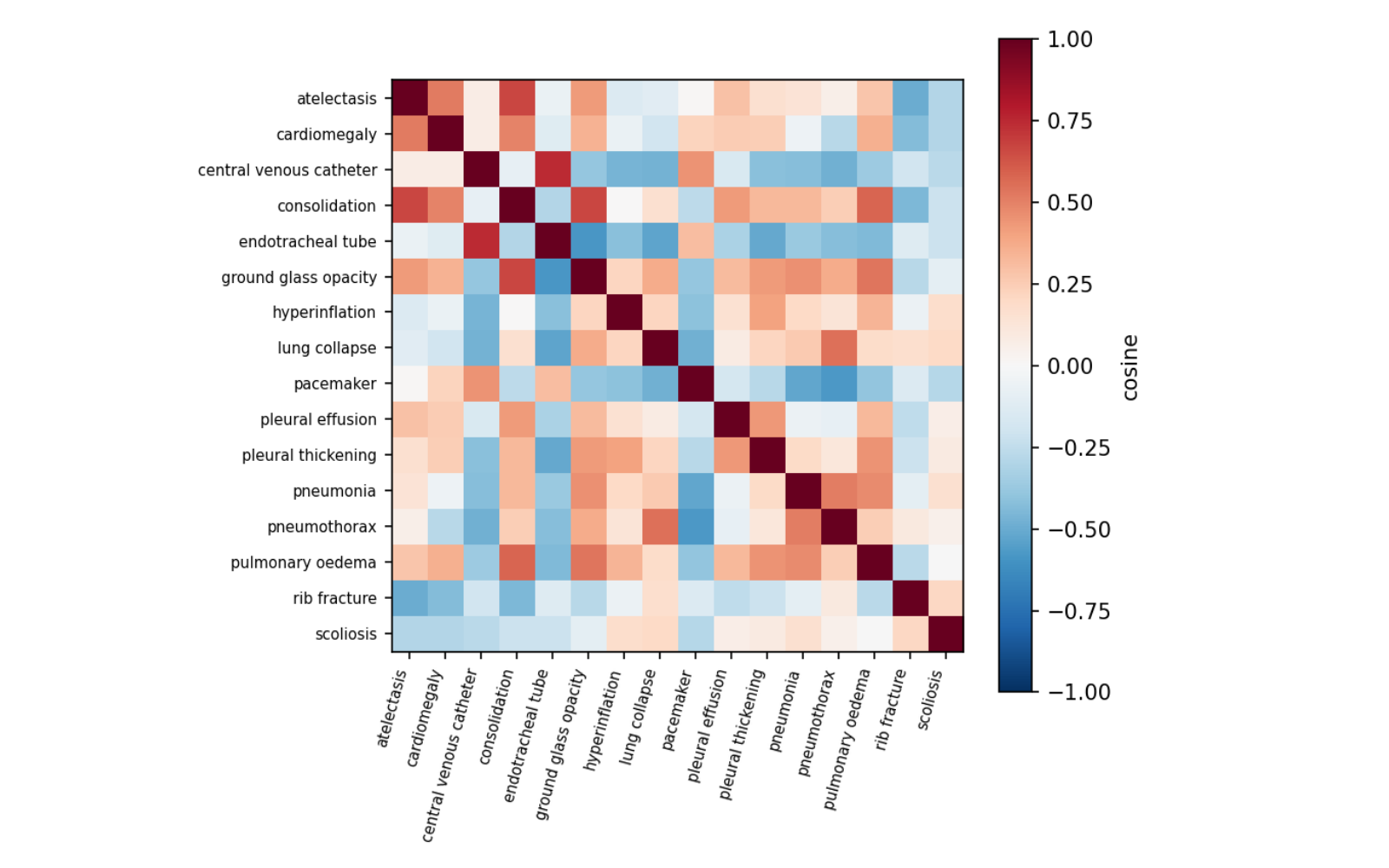}
  \caption{CXR concept cosine similarity.}
\end{figure}

\begin{figure}[htbp]
  \centering
  \includegraphics[width=0.8\linewidth]{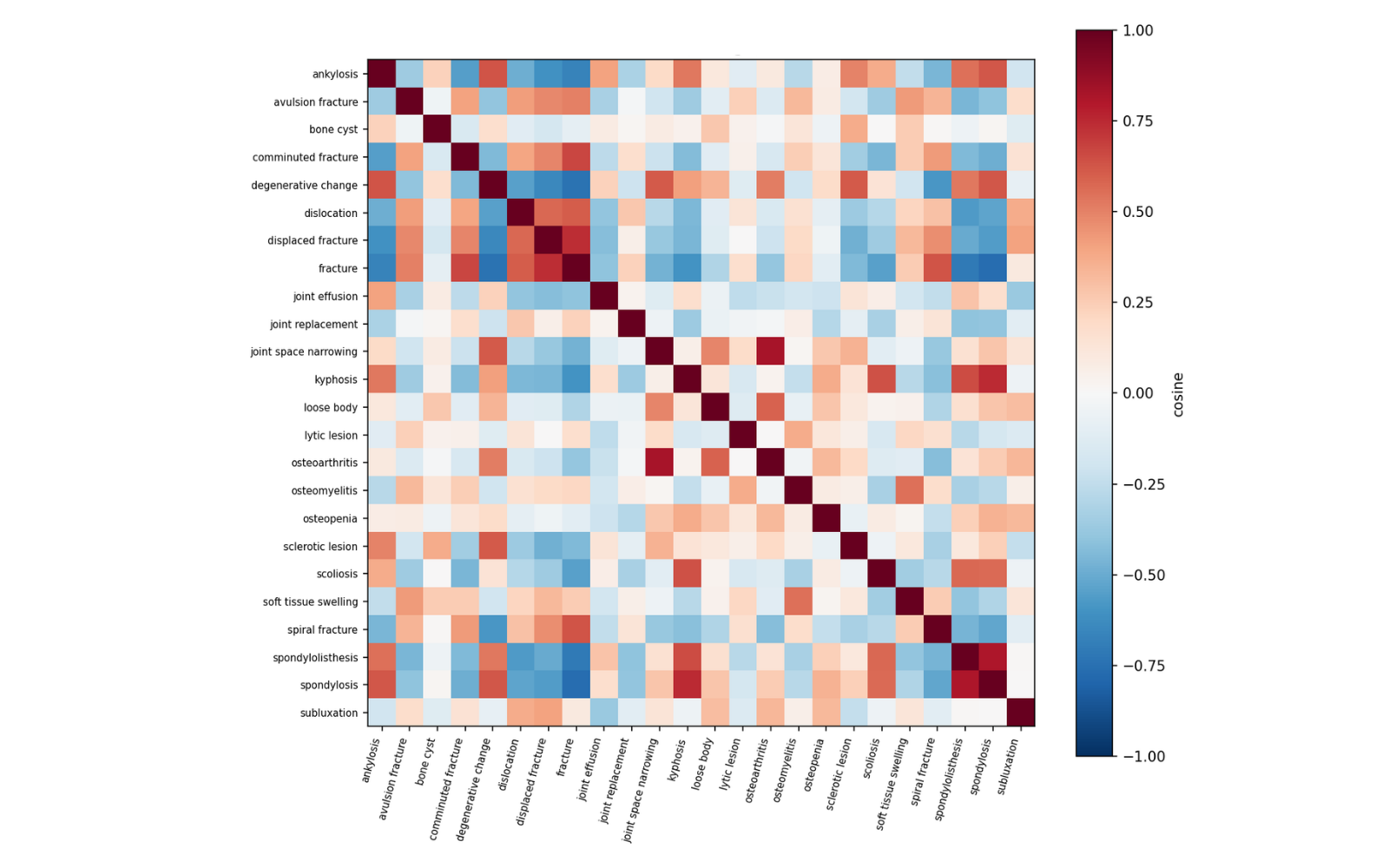}
  \caption{MSK concept cosine similarity.}
\end{figure}

Where directions do align, the alignment tracks clinical relationships rather than arbitrary correlation. In CXR, the overlapping airspace and parenchymal processes group together: atelectasis and consolidation are the most aligned pair (0.77), with lung collapse (0.69) and ground glass opacity (0.67) nearby, while the support devices form their own cluster, the central venous catheter and endotracheal tube aligning at 0.75. In MSK, the degenerative-spine findings cluster tightly (osteoarthritis with joint space narrowing at 0.82, spondylosis with spondylolisthesis at 0.81, kyphosis with spondylosis at 0.75), and the fracture family clusters separately (the generic fracture aligning with displaced fracture at 0.74 and comminuted fracture at 0.67). The two clusters sit on opposite sides of the space: acute fracture directions are most anti-aligned with chronic degenerative ones (fracture versus spondylosis at $-0.76$), mirroring the clinical distinction between traumatic and degenerative disease. The representation therefore organises findings the way a radiologist would group them.

Beyond confirming that each finding is decodable, we test whether the model actually uses these directions rather than merely carrying them. For samples negative for a concept, the probe direction $d$ is added to the residual stream during a forward pass as $\Delta = \text{strength}\cdot s\cdot d$ at every token position, with $s$ the mean residual-stream norm so that strength is comparable across layers, swept over $\{-1.0, 0.0, +1.0, +2.0\}$. This follows additive activation steering \citep{Turner2023,Rimsky2024} along an inference-time intervention direction \citep{LiITI2023}, and the effect is read as the shift in the concept's first-subword token logit relative to the unsteered baseline. We classify each concept vector by the control it affords, using suppression ($\Delta$logit $< 0$ at strength $-1.0$) and induction ($\Delta$logit $> 0$ at both $+1.0$ and $+2.0$) as the two tests:

\begin{itemize}[leftmargin=1.4em]
  \item \textbf{Bidirectional}: subtracting suppresses the finding and adding induces it; the model both reads the direction and writes to it, so it is a usable handle for monitoring and for controlled intervention in either direction.
  \item \textbf{Subtractive}: subtracting reliably suppresses the finding, but adding does not sustain its induction; dependable for detection and suppression, not for induction.
  \item \textbf{Additive}: adding induces the finding, but subtracting does not suppress it; usable for eliciting or amplifying the finding, not for detection or suppression.
  \item \textbf{None}: neither direction produces the expected effect.
\end{itemize}

Tables~\ref{tab:cxr-steer} and~\ref{tab:msk-steer} report the mean $\Delta$logit at each strength for every concept, with its classification.

\begin{table}[H]
\centering
\caption{CXR concept steering: mean $\Delta$logit at each intervention strength.}
\label{tab:cxr-steer}
\small
\begin{adjustbox}{max width=\textwidth}
\begin{tabular}{@{}lcccc@{}}
\toprule
\textbf{Concept} & \textbf{$\Delta$logit @ $-1.0$} & \textbf{$\Delta$logit @ $+1.0$} & \textbf{$\Delta$logit @ $+2.0$} & \textbf{Behaviour} \\
\midrule
pacemaker & $-2.18$ & $+5.86$ & $+7.61$ & bidirectional \\
hyperinflation & $-3.49$ & $+4.21$ & $+4.71$ & bidirectional \\
pulmonary oedema & $-2.72$ & $+3.12$ & $+3.49$ & bidirectional \\
cardiomegaly & $-4.32$ & $+1.36$ & $+1.39$ & bidirectional \\
scoliosis & $-6.10$ & $+2.66$ & $+1.31$ & bidirectional \\
pleural thickening & $-2.35$ & $+1.64$ & $+1.13$ & bidirectional \\
rib fracture & $-3.59$ & $+0.71$ & $+0.95$ & bidirectional \\
endotracheal tube & $-1.02$ & $+0.91$ & $+0.50$ & bidirectional \\
pneumonia & $-3.82$ & $+1.09$ & $-0.57$ & subtractive \\
consolidation & $-2.22$ & $-1.88$ & $-1.48$ & subtractive \\
ground glass opacity & $-1.39$ & $-1.24$ & $-1.77$ & subtractive \\
subcutaneous emphysema & $-3.55$ & $-0.53$ & $-1.87$ & subtractive \\
central venous catheter & $-0.60$ & $-2.13$ & $-2.63$ & subtractive \\
pneumothorax & $-3.65$ & $-0.60$ & $-2.64$ & subtractive \\
lung collapse & $-3.22$ & $-2.56$ & $-4.54$ & subtractive \\
pleural effusion & $-2.01$ & $-3.20$ & $-4.95$ & subtractive \\
atelectasis & $-4.44$ & $-7.44$ & $-7.63$ & subtractive \\
\bottomrule
\end{tabular}
\end{adjustbox}
\end{table}

\begin{longtable}{@{}lcccc@{}}
\caption{MSK concept steering: mean $\Delta$logit at each intervention strength.}
\label{tab:msk-steer}\\
\toprule
\textbf{Concept} & \textbf{$\Delta$logit @ $-1.0$} & \textbf{$\Delta$logit @ $+1.0$} & \textbf{$\Delta$logit @ $+2.0$} & \textbf{Behaviour} \\
\midrule
\endfirsthead
\toprule
\textbf{Concept} & \textbf{$\Delta$logit @ $-1.0$} & \textbf{$\Delta$logit @ $+1.0$} & \textbf{$\Delta$logit @ $+2.0$} & \textbf{Behaviour} \\
\midrule
\endhead
\bottomrule
\endfoot
soft tissue swelling & $-10.07$ & $+11.41$ & $+11.42$ & bidirectional \\
kyphosis & $-5.42$ & $+6.36$ & $+6.28$ & bidirectional \\
degenerative change & $-4.88$ & $+5.14$ & $+4.88$ & bidirectional \\
comminuted fracture & $-6.22$ & $+4.47$ & $+4.55$ & bidirectional \\
osteoarthritis & $-5.67$ & $+2.68$ & $+2.33$ & bidirectional \\
joint space narrowing & $-4.35$ & $+2.33$ & $+2.32$ & bidirectional \\
avulsion fracture & $-9.08$ & $+2.26$ & $+2.32$ & bidirectional \\
dislocation & $-8.28$ & $+2.00$ & $+2.03$ & bidirectional \\
fracture & $-8.86$ & $+1.54$ & $+1.52$ & bidirectional \\
scoliosis & $-9.55$ & $+0.87$ & $+0.93$ & bidirectional \\
subluxation & $-9.83$ & $+1.08$ & $+0.75$ & bidirectional \\
spondylolisthesis & $-9.56$ & $-0.73$ & $-1.10$ & subtractive \\
spiral fracture & $-5.97$ & $-1.56$ & $-1.57$ & subtractive \\
loose body & $-3.89$ & $-3.75$ & $-3.70$ & subtractive \\
joint replacement & $-7.10$ & $-4.13$ & $-4.16$ & subtractive \\
spondylosis & $-9.06$ & $-4.85$ & $-5.25$ & subtractive \\
displaced fracture & $-7.46$ & $-5.10$ & $-5.39$ & subtractive \\
sclerotic lesion & $-7.80$ & $-5.37$ & $-5.43$ & subtractive \\
joint effusion & $-5.53$ & $-5.47$ & $-5.49$ & subtractive \\
lytic lesion & $-2.80$ & $-5.55$ & $-5.63$ & subtractive \\
osteopenia & $-8.04$ & $-6.57$ & $-6.93$ & subtractive \\
osteomyelitis & $-7.60$ & $-7.12$ & $-7.41$ & subtractive \\
bone cyst & $-4.40$ & $-7.39$ & $-7.44$ & subtractive \\
ankylosis & $-9.91$ & $-9.74$ & $-10.21$ & subtractive \\
\end{longtable}

In both domains subtraction lowers the corresponding logit for every concept, so each direction is at least subtractive; no additive-only or unresponsive cases were observed. Eight of 17 CXR concepts and 11 of 24 MSK concepts are fully bidirectional, including many clinically central findings (pulmonary oedema, cardiomegaly, and rib fracture in CXR, and comminuted fracture, dislocation, osteoarthritis, and a generic fracture in MSK) for which the model both reads the direction and writes to it. The remaining concepts are dependable for detection and suppression; their use for induction should be confirmed against generated content rather than assumed. On naturalistic transcripts the directions also fire preferentially on the tokens where their finding is discussed (mean peak projection ratio of present to absent cases: CXR 1.13$\times$, MSK 1.81$\times$), confirming that they carry genuine signal under deployment-like conditions.

\subsection{Concept Localisation: Attention as a Proxy}
\label{sec:localisation}

Localisation answers \textbf{``where in the image did this answer come from?''} by attributing each generated answer, and each individual answer word, back to spatial regions of the input. The engine offers several attribution methods that all produce the same artefact, a heatmap overlaid on the image, but derive it from different internal signals. Offering several is deliberate: gradient-based and attention-based attributions have different failure modes, and agreement between them is more trustworthy than any single map.

\begin{itemize}[leftmargin=1.4em]
  \item \textbf{GradCAM (answer-level)}: A gradient-based saliency method in the spirit of GradCAM \citep{Selvaraju2017}, adapted for HR1.5, attributes the model's answer to the visual features that most influenced it and renders them as a heatmap over the image. The heatmap is \textbf{question-sensitive}: asking ``Is there a pneumothorax?'' vs ``Are there any rib fractures?'' on the same chest X-ray produces different maps focused on the relevant anatomical regions.
  \item \textbf{Per-token gradient attribution}: Because the answer is generated word by word, the same attribution can be computed for a single selected word rather than the whole answer, isolating the region responsible for ``effusion'' separately from the region responsible for ``left''.
  \item \textbf{Attention rollout} \citep{Abnar2020}: Composes the model's attention across layers to approximate the cumulative flow of information that single-layer attention misses.
  \item \textbf{Layer-wise Relevance Propagation (LRP)} \citep{Chefer2021}: Propagates relevance backward from the answer to the input image regions, giving a more principled attribution than raw attention weights.
  \item \textbf{Cross-modal attention}: Reads out, directly from the model's attention, which parts of the image each part of the answer drew on, requiring no gradients.
\end{itemize}

\textbf{Mapping back to the image}: Each method produces a value per image region, which is arranged into a spatial map and overlaid on the image. To keep the overlay faithful, regions that do not correspond to image content are excluded before the map is formed. The map can also be distilled into a set of discrete regions of interest, the contiguous areas of strongest attribution.

\textbf{Attention as a proxy}: It is well established that attention weights are not, by themselves, faithful explanations \citep{Jain2019}, though counter arguments show they are not meaningless either \citep{Wiegreffe2019}. These maps should not be read as a direct delineation of pathology: they indicate the regions the model attended to in forming its answer, not a segmentation of the finding itself. This distinction shapes how their quality should be measured. Overlap metrics such as IoU or DICE, which reward pixel-accurate agreement with a reference mask, are the wrong yardstick for an attention map: the model can attend to the correct anatomical region without its attribution tracing the lesion's exact boundary. The appropriate measure is a \textbf{hit rate}: whether the attended region falls on the clinically relevant area the model should have considered when forming its view. We therefore adopt hit rate as a fidelity score for attended regions, and report it on two localisation datasets below.

We evaluate localisation on two public datasets that provide reference regions: FracAtlas \citep{Abedeen2023}, a musculoskeletal radiograph dataset with fracture bounding boxes, for MSK; and MS-CXR \citep{Boecking2022}, a chest X-ray phrase-grounding dataset with radiologist-drawn bounding boxes, for CXR. A case counts as a hit when the model's attended region falls within the reference region for the finding. We report the localisation hit rate alongside answer accuracy on the same cases (Table~\ref{tab:localisation}).

\begin{table}[H]
\centering
\caption{Localisation hit rate and answer accuracy on FracAtlas (MSK) and MS-CXR (CXR).}
\label{tab:localisation}
\begin{tabular}{@{}lccc@{}}
\toprule
\textbf{Dataset} & \textbf{Modality} & \textbf{Answer accuracy} & \textbf{Localisation hit rate} \\
\midrule
FracAtlas \citep{Abedeen2023} & MSK & \textbf{90.2\%} & 78.7\% \\
MS-CXR \citep{Boecking2022} & CXR & 61.2\% & \textbf{90.2\%} \\
\bottomrule
\end{tabular}
\end{table}

The model attends to the clinically relevant region in the large majority of cases in both modalities (78.7\% MSK, 90.2\% CXR). Notably, on MS-CXR the attended region is correct 90.2\% of the time even though answer accuracy on the same cases is lower (61.2\%), indicating that the model looks in the right place even when it does not arrive at the expected answer, evidence that the localisation reflects genuine visual grounding rather than post-hoc rationalisation of the answer (Figure~4.3).

\begin{figure}[htbp]
  \centering
  \includegraphics[height=0.8\textheight,width=\linewidth,keepaspectratio]{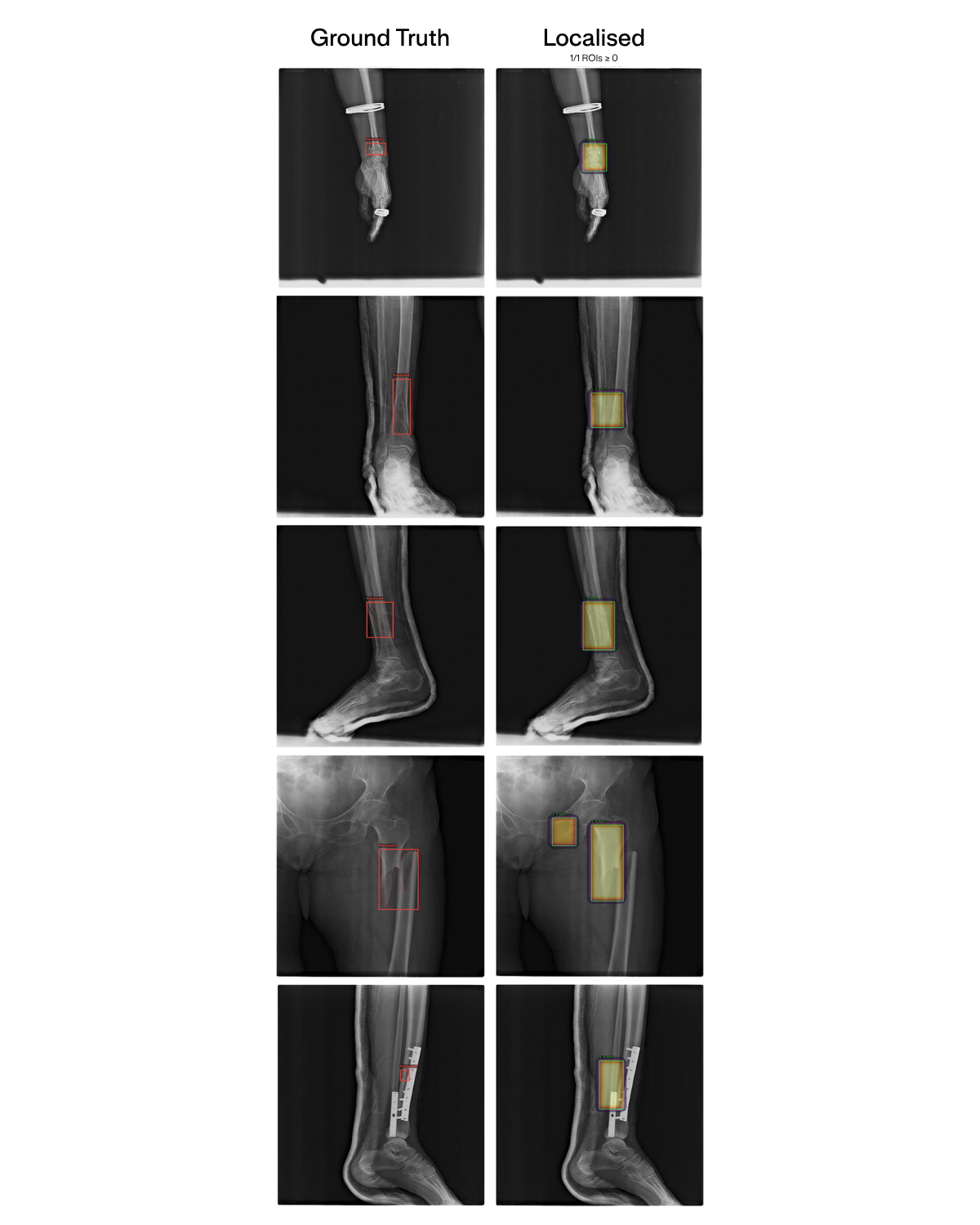}
  \caption{FracAtlas localisation: ground-truth fracture regions (left of each pair) versus the model's attended regions (right) across a sample of cases. MS-CXR examples cannot be shown here owing to dataset licensing restrictions, but the localisation behaviour is analogous and is deducible from the hit-rate results above.}
\end{figure}

These two gaps map onto the classical taxonomy of radiological interpretation error established through eye-tracking studies of visual search \citep{Kundel1978}, which distinguishes \textit{search} errors (the finding is never fixated), \textit{recognition}/perception errors (the region is fixated but the abnormality is not perceived, a category tied in human readers to insufficient dwell time), and \textit{cognition}/decision errors (the abnormality is perceived but misinterpreted). The MSK gap, where the attended region more often falls outside the reference, is the analogue of a \textbf{search error} --- the model's attention does not reach the relevant area. The CXR gap, where the region is attended but the answer is still wrong, is the analogue of a \textbf{cognition error} --- the model looks in the right place but draws the wrong conclusion. The intermediate, dwell-dependent recognition error has no natural counterpart here: inference completes in a single rapid forward pass, so dwell time is not a meaningful quantity for these systems.

\subsection{Confidence and Certainty about the Response}
\label{sec:confidence}

Generative models, including large language models and vision language models, do not natively expose any signal of how confident they are in a given output. In high-stakes settings, where each output should be reviewed for correctness and clinical impact, this absence is a real barrier to adoption: a reader has no internal cue for which answers deserve closer scrutiny. Emerging approaches aim to derive a meaningful confidence estimate from the model's own computation, but they remain early in their development and depend on substantial stabilisation and post-hoc steps such as tuning/calibration before a model's reported confidence can be aligned with its actual reliability. With HR1.5 we explore this direction, with the aim of giving our users a more meaningful signal alongside each answer. The method described below is an early implementation: it already surfaces useful, discriminative signals, though it is not yet tuned/calibrated.

The confidence feature answers \textbf{``how much should this answer be trusted?''} by converting three internal signals (Table~\ref{tab:confidence-signals}), all read at the \textbf{final generated token}, into a single uncertainty value, with confidence defined as its complement \citep{Galil2023}; in the spirit of \citet{Kadavath2022}, showing a model's activations carry information about its own reliability:

\begin{table}[H]
\centering
\caption{Internal signals combined into the confidence (uncertainty) feature.}
\label{tab:confidence-signals}
\small
\begin{tabular}{@{}p{4.4cm}p{9.2cm}@{}}
\toprule
\textbf{Signal} & \textbf{Description} \\
\midrule
Attention entropy & Shannon entropy of the last token's attention distribution, normalised by \texttt{log(sequence\_length)} to $[0,1]$; diffuse attention indicates the model has not settled on what to attend to. \\
Inter-head disagreement & Variance across attention heads of their attention distributions for that token; when heads disagree the aggregate is less reliable. \\
Representation-trajectory instability & Coefficient of variation of the L2 norm of the token's hidden state across layers; spikes or collapses indicate an unsettled representation. \\
\bottomrule
\end{tabular}
\end{table}

These are combined as a weighted sum $U = w_e\,H_{\text{attn}} + w_d\,D_{\text{head}} + w_{cv}\,\mathrm{CV}$, inverted to $\text{confidence} = 1 - U$ (clipped to $[0,1]$), and bucketed for display as \textbf{HIGH / MEDIUM / LOW}. This is a \textit{relative} trust indicator, not a calibrated probability of correctness.

\textbf{Why the raw score is biased by attention sinks}: Transformer models exhibit \textbf{attention sinks} \citep{Xiao2024,Sun2026}, a small number of tokens (often positional or low-content) that attract a disproportionate share of attention mass regardless of semantics, tied to \textbf{massive activation spikes} in a few hidden channels. A sink concentrates attention mass on one or two tokens, making the distribution look \textit{sharp} (low entropy), which the score above reads as \textit{high confidence}, but the sharpness is a structural artefact of the architecture, not evidence the model is genuinely certain about \textit{this} image and \textit{this} question. Left uncorrected, sinks therefore \textbf{deflate entropy and inflate confidence}, and do so most where the model is least engaged with the content.

\textbf{Sink detection and correction}: The engine corrects for this in three steps:

\begin{enumerate}[leftmargin=1.6em]
  \item \textbf{Detect}: for each candidate key position the received attention mass is averaged over batch, heads, and query positions; any position exceeding a fixed threshold is flagged as a sink (the same scan reports per-head sink ratios).
  \item \textbf{Mask and renormalise}: the sink columns are removed from the last token's attention row and the remainder renormalised to a proper distribution (\texttt{p\~{} = p[non-sink] / $\Sigma$ p[non-sink]}); entropy and head disagreement are recomputed on this reduced support. The representation-trajectory term is intrinsic to the hidden state and left unchanged.
  \item \textbf{Recompute and report inflation}: the composite uncertainty is recalculated, and the \textbf{confidence inflation} $\text{inflation} = \max(0,\, (1 - U_{\text{raw}}) - (1 - U_{\text{corrected}}))$ is reported explicitly. By construction it is non-negative: the correction can only \textit{reveal} hidden uncertainty, never manufacture confidence.
\end{enumerate}

Rather than silently replacing the number, the engine surfaces \textbf{both} values and the gap between them, for example, \textit{``Raw: HIGH $\rightarrow$ Corrected: MEDIUM (sinks inflate by 0.23)''}, so a reader can see \textit{when} the model's apparent certainty was an artefact. This is a post-hoc, signal-level correction complementary to, not a substitute for, formal probability calibration such as temperature scaling on held-out data \citep{Guo2017}.

\textbf{Worked example, see figure 4.4}: For the chest X-ray below, asked \textit{``Describe the findings in this chest x-ray,''} HR1.5 responded with ``The chest x-ray shows a 3.5cm mass in the left upper lobe, which is suspicious for malignancy. There are no other pulmonary nodules or masses identified. No pleural effusion is present. The cardiac size and mediastinal contours appear normal.''

\begin{figure}[htbp]
  \centering
  \includegraphics[width=0.8\linewidth]{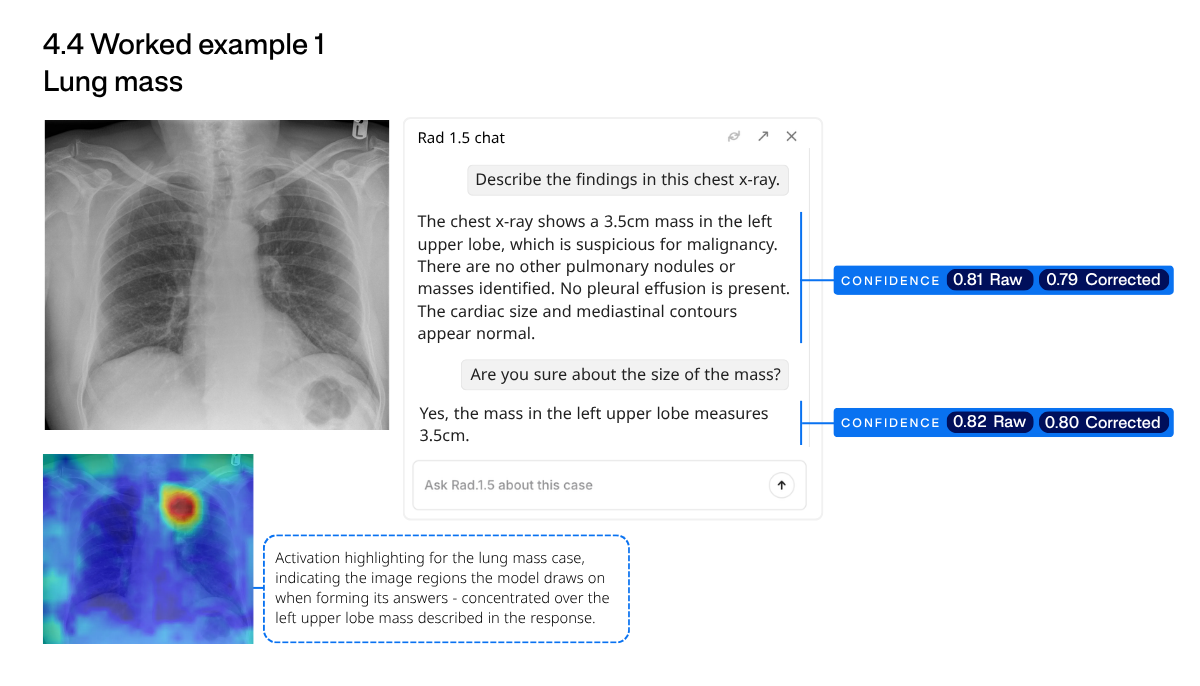}
  \caption{Left apical lung mass case. Case courtesy of Dai Roberts, Radiopaedia.org (case \href{https://radiopaedia.org/cases/80445}{80445}), used under licence.}
\end{figure}

The engine reports a raw confidence of \textbf{0.81}. Detecting and correcting for attention sinks lowers this to \textbf{0.79}, an inflation of 0.03. A small portion of the apparent certainty was a structural artefact rather than genuine engagement with the image and question. The modest gap leaves the overall trust assessment unchanged, while making explicit that the corrected figure is the more faithful one.

On the follow-up question \textit{``Are you sure about the size of the mass?''}, the model answered \textit{``Yes, the mass in the left upper lobe measures 3.5cm,''} with a raw confidence of \textbf{0.82} corrected to \textbf{0.80}. The confidence holds steady under direct challenge, and the sink correction again accounts for only a small (0.03) inflation, indicating the model's certainty about the measurement is genuine rather than an attention artefact.

\textit{Activation highlighting for the lung mass case indicates the image regions the model draws on when forming its answers, concentrated over the left upper lobe mass described in the response.}

\textbf{Worked example, shown in figure 4.5, a multi-turn conversation}: The exchange below is on a Weber B ankle fracture case. Each turn shows the question, the model's answer, and the raw and sink-corrected confidence.

\begin{figure}[htbp]
  \centering
  \includegraphics[width=0.8\linewidth]{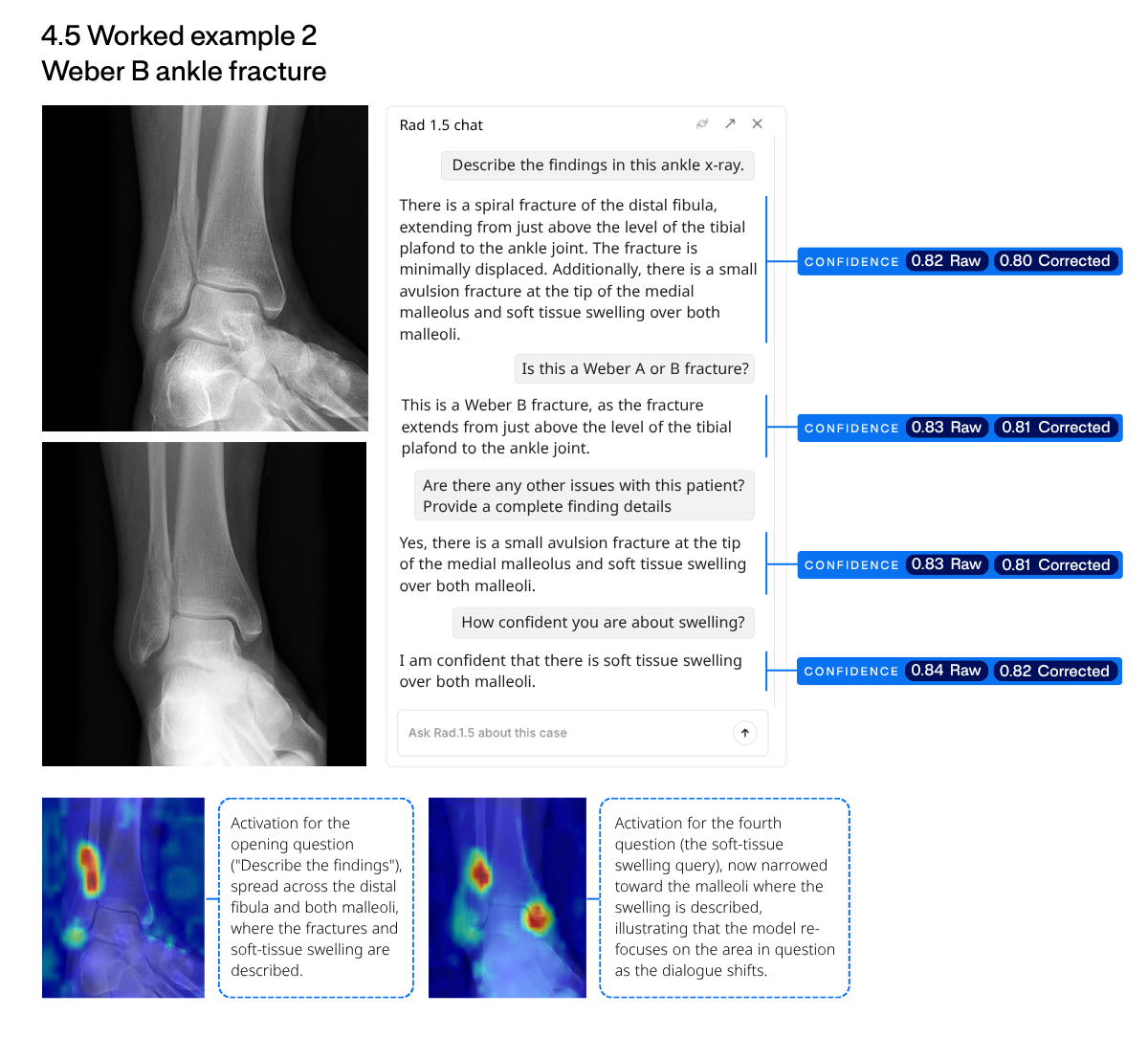}
  \caption{Ankle Weber B fracture. Case courtesy of Frank Gaillard, Radiopaedia.org (case \href{https://radiopaedia.org/cases/7480}{7480}), used under licence.}
\end{figure}

Across the conversation the corrected confidence stays in a narrow band (0.80--0.82) with a consistent, small sink inflation ($\sim$0.02), showing that the model's certainty is stable as the dialogue moves from description to classification to a full report, and does not spuriously spike. The activation highlighting shows where in the image the model draws this information from, and how that focus shifts as the question changes between the opening description and the final challenge. Activation for the opening question (``Describe the findings'') is spread across the distal fibula and both malleoli, where the fractures and soft-tissue swelling are described; activation for the fourth question (the soft-tissue swelling query) is narrowed toward the malleoli where the swelling is described, illustrating that the model re-focuses on the area in question as the dialogue shifts.

\textbf{Worked example, shown in figure 4.6, an out-of-distribution input}: To probe how the confidence signal behaves on atypical inputs, we presented the first medical radiograph ever taken (Röntgen's 1895 X-ray of a hand) with a deliberately off-domain question.

\begin{figure}[htbp]
  \centering
  \includegraphics[width=\linewidth]{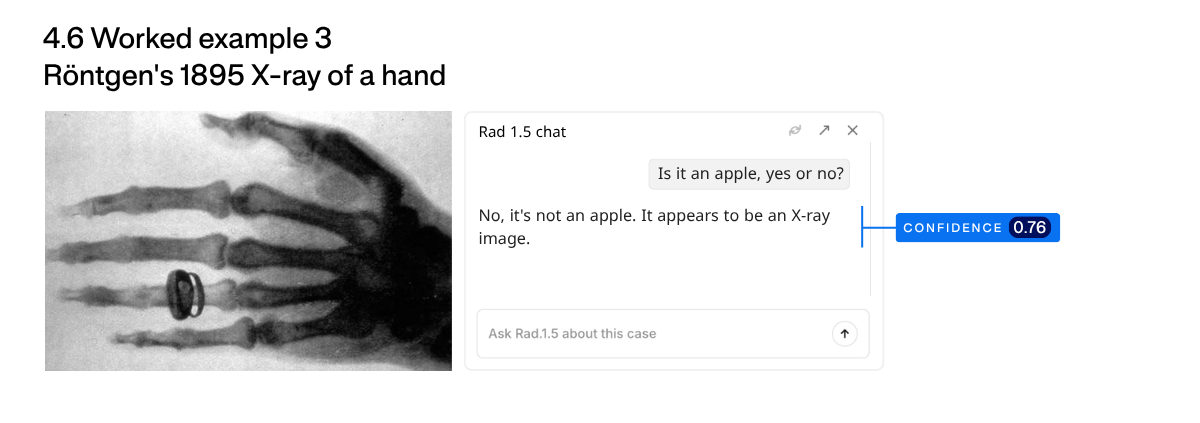}
  \caption{Querying with the historical first Röntgen's 1895 X-ray of a hand. Underlying radiograph: \href{https://commons.wikimedia.org/wiki/File:X-ray_by_Wilhelm_R\%C3\%B6ntgen_of_Albert_von_K\%C3\%B6lliker\%27s_hand_-_18960123-02.jpg}{Wikimedia Commons} (Wilhelm Röntgen, X-ray of Albert von Kölliker's hand, 1896).}
\end{figure}

The answer is correct, but the confidence falls to 0.76, noticeably below the 0.80+ reported on the in-distribution clinical cases above. The lower figure tracks the unusual, low-contrast, historically degraded image and the off-domain question: the confidence signal registers that this input sits outside the model's familiar distribution, which is exactly the behaviour a trustworthy uncertainty estimate should show.

While these results make this a promising direction for ascertaining confidence, the approach remains at an early stage and will require further development before it can serve as a consistent, interpretable indicator of confidence. We are optimistic that, with further stabilisation and calibration, these signals can be made substantially more meaningful and reliable, and refining them is an area we intend to focus on in future work.

\subsection{Behaviour on Out-of-Distribution Inputs}

A model's response to inputs it could never have seen in training is as revealing as its performance on genuine studies. To probe this, we presented a synthetic chest X-ray generated with the GPT-5.5 image API from the prompt \textit{``generate a realistic looking x-ray of an apple in a human chest''}, an image that looks radiographic but contains an object with no counterpart in any real study (figure 4.7).

The model accepts the synthetic image as a genuine radiograph and, when reporting findings, maps the unfamiliar object onto the nearest concept it has learnt (a left breast prosthesis) holding that interpretation with high confidence even when directly challenged. This exposes a real limitation: on concepts entirely outside its training distribution, the model is incorrect. Importantly, though, the error is a \textit{reasonable} one: the synthetically generated apple has overlapping imaging characteristics with breast prosthesis or implant. The output is plausible given the model training data, rather than being outright nonsensical.

This points to a limitation that no confidence signal can fully remove: when an out-of-distribution input maps plausibly onto a learnt concept, as the apple does onto a prosthesis, the model can be confidently wrong, so a high value alone does not guarantee correctness. As the Röntgen case above shows, the signal can sometimes register an unfamiliar input as less certain, though not yet reliably. In our own system we already gate on this, rejecting inputs identified as out-of-distribution so that unreliable cases are held back rather than reported. Calibrating these confidence signals and measuring their reliability quantitatively remains future work; what we present here is an early, qualitative result.

\begin{figure}[htbp]
  \centering
  \includegraphics[width=\linewidth]{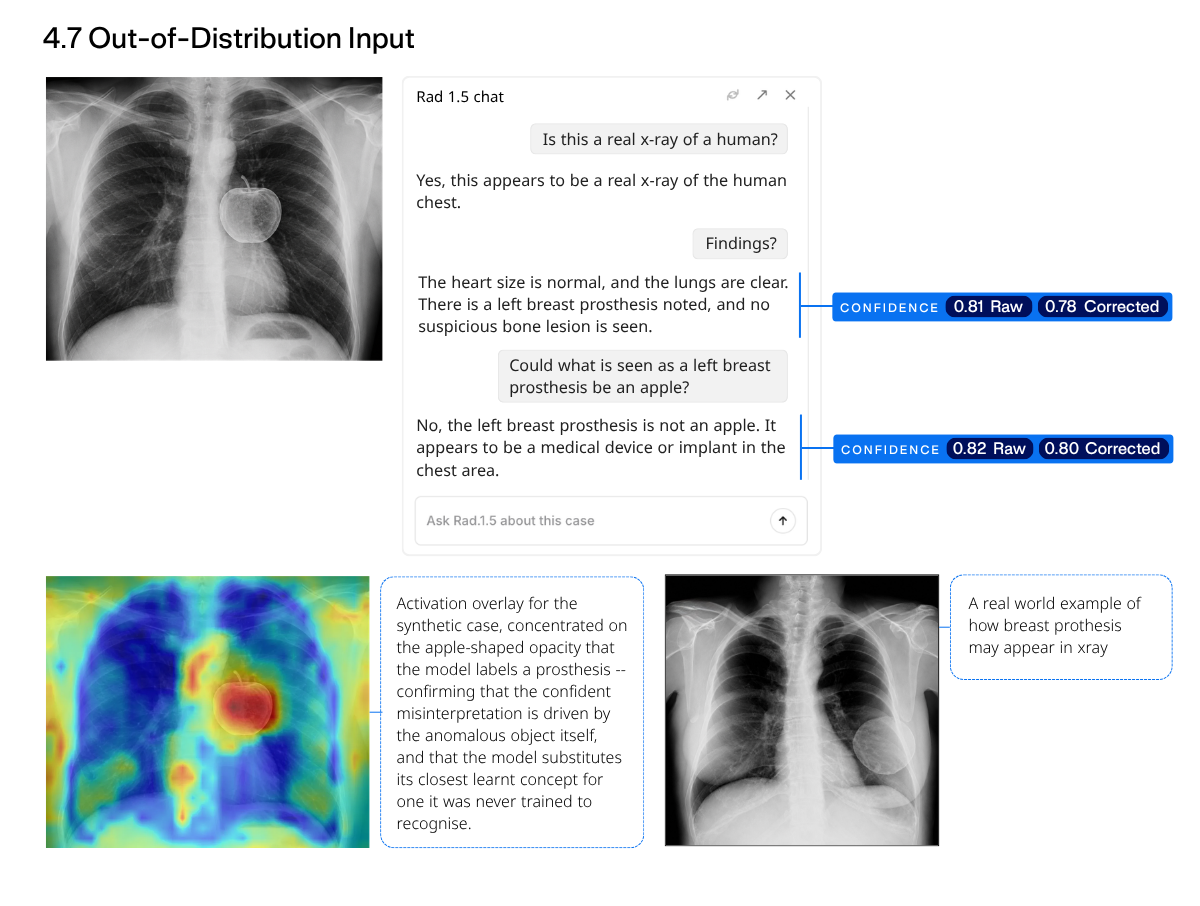}
  \caption{Query with a synthetically generated x-ray image. The activation overlay is concentrated on the apple-shaped opacity that the model labels a prosthesis, confirming that the confident misinterpretation is driven by the anomalous object itself, and that the model substitutes its closest learnt concept for one it was never trained to recognise.}
\end{figure}
\section{Discussion}
\label{sec:discussion}

Considered in aggregate, the evaluations support a consistent conclusion: HR1.5 attains its largest advantages on the tasks of greatest clinical consequence. On our internal, held-out mock simulations of the FRCR examinations, which reconstruct the assessments through which the Royal College of Radiologists certifies practising radiologists and serve as the closest available proxy in this study to authentic diagnostic reporting, HR1.5 and its agentic configuration HR1.5+ are the only systems to meet the passing standard, satisfying the Angoff-referenced threshold of the current 2B Short Case format. Every other general-purpose and medical-domain model evaluated falls substantially below it. Because this benchmark most directly approximates the clinical reporting task, the magnitude of this margin is of particular practical significance. We note that these examinations are internally constructed simulations rather than the official RCR sittings, which are not publicly available (Section~\ref{sec:frcr}).

This advantage generalises beyond the examination setting. On closed-format clinical questions and on the multi-body-part internal evaluation, HR1.5 attains the highest accuracy across diagnosis, finding presence, and normal/abnormal classification, and across chest, musculoskeletal, and other anatomical regions, indicating that the improvement is not specific to a single modality or question type.

We are equally explicit about the settings in which HR1.5 does not lead. On a few open-ended chest-radiograph report-generation benchmarks, frontier general-purpose models and chest-specialised systems achieve higher values on individual metrics, and the agentic HR1.5+ configuration does not uniformly improve upon HR1.5, most notably under multi-turn dialogue evaluation. We attribute this in part to genuine headroom that remains to be addressed, and in part to the metric limitations characterised in Section~\ref{sec:evaluation}, in which surface- and entity-level scores reward lexical form over clinical correctness. Neither consideration qualifies the principal result: on the task that most closely resembles, and is most likely to inform, routine radiological practice (examination-grade reporting) HR1.5 outperforms all evaluated systems by a substantial margin.

\section{Future Work}
\label{sec:future}

\subsection{Towards Better Evaluation: The C6 Framework}

Section~\ref{sec:evaluation} established why automated metrics diverge from clinical utility: conventional report-generation metrics track wording far more than clinical correctness, and even the Findings-Diagnosis score we adopt as the primary measure, with the Judge Findings-Diagnosis variant where automated matching is insufficient, remains a single aggregate. As noted there, it inherits the central limitation of all scalar scores: it can credit a report that misses a subtle finding, can accept a vague but incorrect descriptor as a match, and assigns comparable penalties to errors of very different clinical consequence. A report that identifies the correct diagnosis but omits a minor finding is scored similarly to one that hallucinates a dangerous pathology. A scalar score, however well-aligned, does not indicate \textit{why} a report is adequate or inadequate: whether it identified the clinically important findings, whether it would be safe to act upon, and whether a radiologist would accept it.

Recent work has reached the same conclusion, that single-score metrics are insufficient for clinical text, and has begun to decompose evaluation into interpretable axes. DocLens \citep{DocLens2024} evaluates medical text along completeness, conciseness, and attribution; GREEN \citep{GREEN2024} introduces error-aware evaluation with interpretable explanations; RadCliQ \citep{Yu2023} combines several signals into a clinically grounded composite; and CT-FineBench~\citeyearpar{CTFineBench2026} demonstrates that conventional metrics fail to capture granular diagnostic accuracy in CT report generation. These efforts point toward a consensus: clinically meaningful evaluation requires fine-grained, multi-dimensional assessment rather than a single opaque number.

\begin{figure}[htbp]
  \centering
  \includegraphics[width=0.85\linewidth]{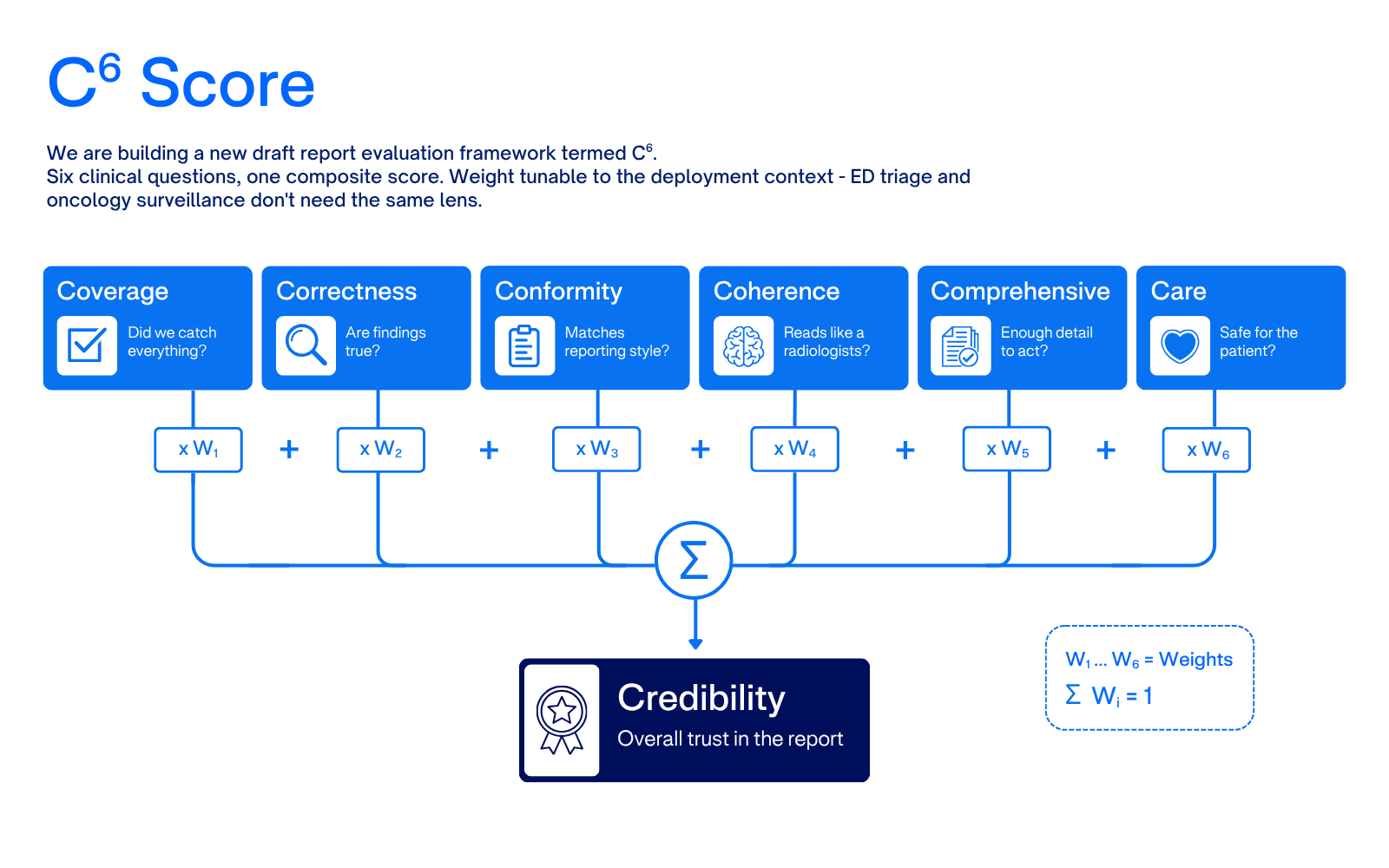}
  \caption{The C6 framework. A draft report is scored along six clinical pillars --- Coverage, Correctness, Comprehensiveness, Care, Coherence, and Conformity --- each yielding a pillar score $p_i$. A weighted sum with deployment-tunable weights $w_i$ produces a single Credibility score, $\sum_i w_i p_i$.}
  \label{fig:c6}
\end{figure}

To this end, we are developing the \textbf{C6 evaluation framework}, which scores a draft report along six interpretable clinical pillars (Table~\ref{tab:c6}) and combines them into a single composite (Figure~\ref{fig:c6}).

\begin{table}[H]
\centering
\caption{The six clinical pillars of the C6 framework.}
\label{tab:c6}
\begin{tabular}{@{}lp{9.6cm}@{}}
\toprule
\textbf{Pillar} & \textbf{Evaluation question} \\
\midrule
\textbf{Coverage} & Are all findings that should be reported identified? (recall) \\
\textbf{Correctness} & Are the reported findings true, with no hallucinated pathology? (precision) \\
\textbf{Comprehensiveness} & Are correctly identified findings described in sufficient detail to act on? \\
\textbf{Care} & What is the patient-safety impact of any errors? \\
\textbf{Coherence} & Is the report clear, internally consistent, and well-communicated? \\
\textbf{Conformity} & Does the reporting style match institutional expectations? \\
\bottomrule
\end{tabular}
\end{table}

The six pillars can be evaluated independently or combined into an overall \textbf{Credibility} score, defined as the weighted sum $\text{Credibility} = \sum_i w_i p_i$ of the per-pillar scores. The decomposition is what makes the framework diagnostic rather than merely summative: it separates failure modes that a scalar metric conflates, distinguishing a model with high Coverage but low Correctness (over-calling) from one with high Correctness but low Comprehensiveness (under-describing). The Care pillar introduces explicit clinical severity weighting, absent from existing metrics, so that safety-critical errors such as a missed pneumothorax are penalised more heavily than benign omissions or stylistic divergence.

The weights $w_i$ are tunable per deployment, reflecting that different clinical settings require different emphases: emergency triage prioritises Care and Coverage, where detecting an acute, actionable finding is paramount, whereas oncology surveillance emphasises Comprehensiveness and Correctness, where precise characterisation and interval comparison are most important.

Underlying all six pillars is a radiology knowledge base that anchors each extracted finding to a stable clinical concept, extending the ontology-based synonym matching and polarity logic already employed by the Findings-Diagnosis score (Section~\ref{sec:quant}). This grounding enables C6 to recognise that ``no airspace opacification'' contradicts ``consolidation'' despite the lexical dissimilarity, and conversely that ``opacity'' and ``consolidation'' are clinically related rather than contradictory, so that related descriptors are not penalised as opposing assertions.

We intend to adopt C6 as the primary evaluation framework for future model iterations and to release it as an open evaluation tool for the broader community, providing a shared, clinically grounded basis for assessing radiology report quality.

\subsection{3D and Volumetric Imaging Support}

HR1.5 is trained exclusively on 2D X-ray modalities. A natural extension is to support 3D volumetric imaging, particularly CT scans, which represent a substantial and growing share of radiology workload. Recent work has demonstrated the feasibility of foundation models for 3D radiology: CT2Rep \citep{CT2Rep2024} introduced automated report generation for 3D medical images, Argus~\citeyearpar{Argus2025} established the first comprehensive benchmark for 3D radiograph report generation, and several 2026 efforts have tackled challenges specific to volumetric data including entity hallucination mitigation, region-grounded report generation, and hierarchical vision-language modelling for efficient volume processing.

Extending HR1.5 to 3D introduces significant architectural and data challenges. Volumetric inputs are orders of magnitude larger than single X-ray images, requiring efficient spatial encoding strategies, whether through hierarchical 2.5D approaches, sparse attention over volume patches, or multi-view aggregation. The training data pipeline must be adapted for DICOM series with slice-level annotations and volumetric spatial relationships.

We plan to address these challenges in future model iterations, with the goal of delivering a unified radiology foundation model capable of interpreting both 2D and 3D imaging modalities within a single architecture.

\section*{Acknowledgements}
\addcontentsline{toc}{section}{Acknowledgements}

We thank the radiologists and clinical collaborators whose expertise informed the design, scoring, and evaluation of HR1.5, and our colleagues across Harrison.ai for their engineering and research contributions. We gratefully acknowledge the open-source community, in particular PyTorch and Hugging Face and their contributors, whose frameworks supported the training, development, and evaluation of HR1.5, and the vLLM project and its contributors, whose tools underpin our serving stack, and Radiopaedia.org and its contributors for the teaching cases used under licence in the qualitative examples.
\clearpage
\addcontentsline{toc}{section}{References}
\bibliographystyle{plainnat}
\bibliography{references}

\end{document}